\documentclass[preprint,12pt]{elsarticle}

\usepackage[utf8]{inputenc}
\usepackage{textgreek}
\usepackage{booktabs, makecell}
\usepackage{multirow}

\usepackage[T1]{fontenc}
\usepackage{newtxtext, newtxmath} 

\usepackage{graphicx}
\usepackage{subfig}
\usepackage{float}
\usepackage{tcolorbox}
\usepackage{algorithm}
\usepackage{algpseudocode}
\usepackage{tikz}
\usetikzlibrary{shapes.geometric, arrows, positioning}

\usepackage{longtable}
\usepackage{tabularx}
\usepackage{rotating}
\usepackage{booktabs}       
\usepackage{multirow}       
\usepackage{threeparttable} 
\usepackage{siunitx}   

\usepackage{amsmath}

\usepackage{newtxtext,newtxmath}
\usepackage{amsfonts}
\usepackage{mathtools}
\usepackage{minted}

\usepackage{listings}
\usepackage{xcolor}

\usepackage{graphicx}

\usepackage{hyperref}
\hypersetup{
    colorlinks=true,
    linkcolor=blue,
    citecolor=blue,
    urlcolor=blue,
    pdftitle={Cost-Effective Fine-Tuning of Financial LLMs},
    pdfauthor={Your Name},
}
\usepackage{cleveref}

\usepackage{xspace}
\usepackage{xcolor}
\usepackage{enumitem}

\journal{Engineering Applications of Artificial Intelligence}

\begin{document}

\begin{frontmatter}

\title{When Financial Fine-tuning Fails: A Three-Level Detectability Analysis of Numerical Hallucination in Domain-Adapted Language Models}

\author[inst1,inst2]{Xiaodong Li\corref{cor1}}
\ead{lixd@cofco.com}

\author[inst3]{Peiwei Liu}
\ead{liupeiwei@gzasc.edu.cn}

\affiliation[inst1]{organization={School of Computer Science},
  addressline={Guangzhou College of Applied Science and Technology},
  city={Guangzhou},
  postcode={511370},
  country={China}}

\affiliation[inst2]{organization={IT Department},
  addressline={COFCO Corporation},
  city={Beijing},
  postcode={100020},
  country={China}}

\affiliation[inst3]{organization={School of Computer Science},
  addressline={Guangzhou College of Applied Science and Technology},
  city={Guangzhou},
  postcode={511370},
  country={China}}
  
\cortext[cor1]{Corresponding author}
\begin{abstract}
Financial large language models are increasingly deployed for summarization of reports and disclosures, where numerical hallucination poses significant practical risks. While prior work often attributes such hallucination to insufficient numerical reasoning, this assumption has not been systematically tested under controlled fine-tuning settings.
In this paper, we conduct a cost-effective, controlled study of numerical hallucination in financial summarization across three model variants: a base instruction-tuned model, a domain language-adapted model (FT-A), and a numeracy-enhanced domain model (FT-A+B+C). We introduce a three-level detectability taxonomy distinguishing between overt hallucination (currency-denominated fabrication), covert-explicit hallucination (professional-convention numbers), and covert-implicit hallucination (ungrounded quantitative claims).
Our results reveal that domain fine-tuning substantially degrades numerical restraint at all detectability levels. While the Base model maintains near-zero hallucination rates (5.4\%), FT-A exhibits 82.5\% overt hallucination and FT-A+B+C reaches 98\%. Contrary to intuition, numeracy supervision amplifies rather than mitigates hallucination across all levels. We identify template injection---the insertion of memorized canonical values regardless of input content---as a primary hallucination mechanism in fine-tuned models.
These findings demonstrate that numerical hallucination in financial summarization is driven by the degradation of numerical restraint through domain adaptation, not by insufficient numerical reasoning. We recommend that evaluation protocols assess hallucination across all detectability levels and that deployment practices include explicit mechanisms for grounding-aware generation or abstention.
\end{abstract}

\begin{keyword}
Financial NLP \sep Large Language Models \sep Hallucination \sep Evaluation Framework \sep Domain Adaptation \sep QLoRA
\end{keyword}

\end{frontmatter}


\section{Introduction}
\label{sec:introduction}

\subsection{Background and Motivation}

Large language models (LLMs) have demonstrated remarkable capabilities across diverse natural language processing tasks, leading to their rapid adoption in specialized domains including finance, healthcare, and legal applications\citep{ji2023survey,bubeck2023sparks}. In financial contexts, LLMs are increasingly deployed as core components of expert systems for document summarization, risk assessment, earnings analysis, and investment decision support. The appeal is clear: these models can process vast volumes of financial text and generate professional-quality summaries that would otherwise require substantial human effort.

However, the reliability of LLM-generated content in high-stakes financial applications remains a critical concern. Financial documents demand numerical precision---a misrepresented revenue figure, fabricated growth rate, or invented cash flow statement can mislead investment decisions, violate regulatory requirements, and erode stakeholder trust. Unlike creative writing or general-purpose summarization, financial summarization operates under strict factual constraints where even stylistically appropriate outputs may contain dangerous fabrications.

A common assumption in the deployment of financial language models is that domain-specific fine-tuning improves reliability alongside fluency. Models fine-tuned on financial corpora exhibit improved terminology usage, professional tone, and structural conformity\cite{araci2019finbert, chen2023finma, wu2023bloomberggpt}. An extension of this assumption holds that numeracy-enhanced training---incorporating arithmetic supervision and numerical reasoning tasks---should further reduce numerical errors by improving the model's computational competence.

This paper challenges both assumptions through systematic empirical analysis.

\subsection{Problem Statement}

We investigate a fundamental question: \textbf{Does domain fine-tuning improve or degrade numerical reliability in financial summarization?}

This question is motivated by an apparent paradox observed in preliminary experiments. Models fine-tuned on financial data produce outputs that appear more professional and numerically fluent than their base counterparts. They readily generate revenue figures, growth rates, and cash flow statements in appropriate formats. Yet closer inspection reveals that many of these numerical expressions are entirely fabricated---invented values that appear nowhere in the source text.

We hypothesize that this paradox arises from a conflation of two distinct capabilities:

\begin{itemize}
    \item \textbf{Numerical Competence: }The ability to perform correct numerical computations when explicitly required
    \item \textbf{Numerical Discipline: }Numerical Discipline is operationally defined as a model's tendency to abstain from generating absolute numerical quantities when the input provides no such grounding. It is measured directly as the complement of the L1 hallucination rate under S0 conditions: a model with high numerical discipline produces responses free of fabricated absolute values when presented with S0 inputs.
\end{itemize}

Domain fine-tuning may improve competence while destroying discipline, producing models that fabricate numerical content more fluently and more frequently. We term this divergence the \textbf{restraint gap}.

To test this hypothesis, we require an evaluation framework capable of detecting numerical hallucination across varying levels of subtlety. Existing "strict" metrics focus on overt formatting violations (e.g., malformed currency expressions) but may miss professionally-formatted fabrications. We therefore introduce a three-level detectability taxonomy that enables systematic assessment of hallucination across evaluation stringencies.

\subsection{Contributions}

This paper makes the following contributions:
\begin{enumerate}
    \item \textbf{Methodological Framework: }We propose a three-level detectability taxonomy for numerical hallucination evaluation:
    \begin{itemize}
        \item \textbf{L1 (Overt): }Currency-denominated fabrications detectable via pattern matching (e.g., "USD 1.3 billion")
        
        \item \textbf{L2 (Covert-Explicit): }Professional-convention numbers requiring grounding verification (e.g., "3.3 billion", "growth of 11\%")
        
        \item \textbf{L3 (Covert-Implicit):}Implicit quantitative claims requiring semantic understanding (e.g., "maintained stable debt levels")
    \end{itemize}
    This taxonomy enables systematic assessment of model reliability across evaluation stringencies and reveals blind spots in existing evaluation protocols.
    \item \textbf{Empirical Analysis: }Through controlled experiments on 240 financial summarization samples (160 synthetic, 80 real-world 10-K excerpts) across three model variants (Base, FT-A, FT-A+B+C), we demonstrate that:
        \begin{itemize}
        \item domain fine-tuning substantially degrades numerical restraint at \textbf{all} detectability levels, not merely at harder-to-detect levels
        \item Base models maintain near-zero hallucination (5.4\% L1) while fine-tuned models exhibit catastrophic rates (82.5-90.8\% L1)
        \item Numeracy supervision \textbf{amplifies} rather than mitigates hallucination across all levels
    \end{itemize}
    \item \textbf{Mechanism Identification: }We identify \textbf{template injection}---the insertion of memorized canonical values regardless of input content---as a primary hallucination mechanism. Specific fabricated templates (e.g., "USD 3.3 billion", "operating cash flow of USD 1011 million") appear in up to 37-50\% of fine-tuned model outputs across unrelated inputs, suggesting that domain adaptation induces strong narrative priors that override evidence-based generation.
    \item \textbf{Practical Guidelines: }We provide deployment recommendations for financial expert systems, including:
        \begin{itemize}
        \item Multi-level hallucination assessment protocols
        
        \item Grounding condition detection for input routing
        \item Template injection monitoring and filtering
        \item Decision frameworks for model selection and deployment
    \end{itemize}
\end{enumerate}

\subsection{Key Findings Preview}

Our analysis reveals five critical findings that challenge conventional assumptions about financial language model reliability:

\textbf{Finding 1: }Fine-tuning substantially degrades numerical restraint at all detectability levels. The Base model maintains 5.4\% L1 (Overt) hallucination, while FT-A exhibits 82.5\% and FT-A+B+C reaches 90.8\%. This contradicts any assumption that domain-adapted models appear safe under strict evaluation.

\textbf{Finding 2: }Numeracy supervision amplifies hallucination. Despite improving arithmetic competence under constrained evaluation, FT-A+B+C exhibits higher hallucination rates than FT-A at every detectability level (+7.9 points for L1, +11.6 points for L2, +9.5 points for L3).

\textbf{Finding 3: }Weak numerical grounding (S0 conditions) represents a catastrophic vulnerability. Under inputs lacking absolute numerical values, fine-tuned models exhibit near-universal L1 hallucination (90-100\%), while the Base model maintains complete restraint (0.8\%).

\textbf{Finding 4: }Template injection is pervasive, suggesting a strong reliance on memorized boilerplate over input-grounded generation.

\textbf{Finding 5: }The restraint gap is the core problem. Numerical competence and numerical discipline are orthogonal capabilities; current fine-tuning practices optimize the former while destroying the latter.

\begin{figure}[h]
\centering
\includegraphics[width=0.8\columnwidth]{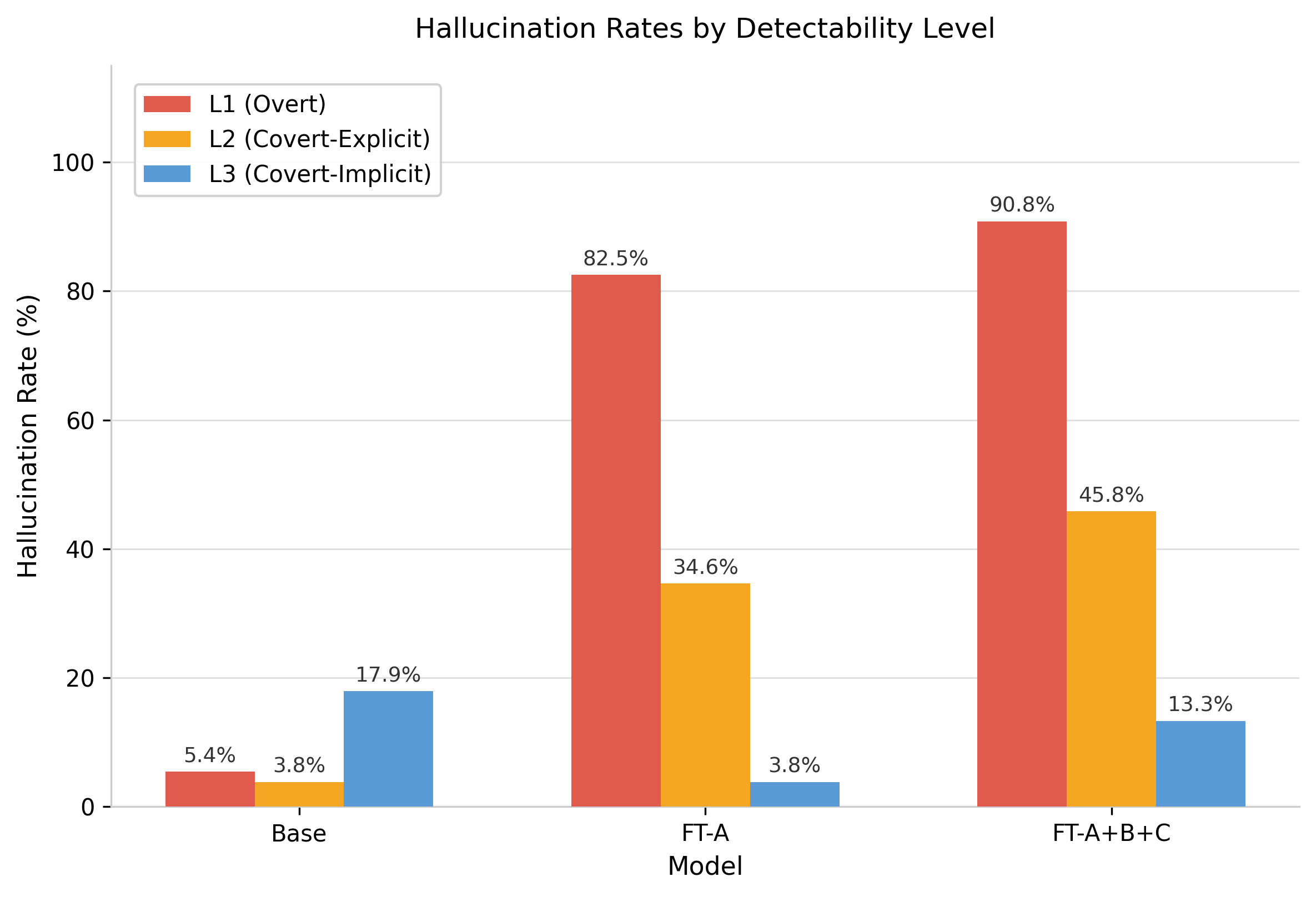}
\caption{Hallucination rates by detectability level across model variants (n=240). L1 (Overt): currency-denominated fabrications; L2 (Covert-Explicit): absolute quantities without currency markers; L3 (Covert-Implicit): ungrounded quantitative claims. Both fine-tuned models exhibit substantially higher hallucination rates than the Base model at all three detectability levels, with FT-A+B+C exceeding FT-A across L1 and L2.}
\label{fig:overall}
\end{figure}

Figure~\ref{fig:overall} previews our central finding: domain 
fine-tuning dramatically increases hallucination at all 
detectability levels.

\section{Related Work}
This study is situated at the intersection of three research areas: hallucination in neural text generation, numerical reasoning and grounding in language models, and domain adaptation for financial language applications. While recent large-scale models have demonstrated impressive emergent 
capabilities across diverse tasks\cite{bubeck2023sparks}, their reliability 
in specialized domains with strict factual requirements remains a concern.

\subsection{Hallucination in Neural Text Generation}
Hallucination, broadly defined as the generation of content not supported by the input or source context, has been extensively studied in summarization, question answering, and dialogue systems. Recent benchmarks such as TruthfulQA \cite{lin2021truthfulqa} have 
revealed that large language models frequently generate false statements, particularly when prompted with questions designed to elicit common misconceptions. Early work characterizes hallucination as fluent yet factually incorrect generation, often arising from model overgeneralization or exposure bias \cite{maynez2020faithfulness,cao2020factual,goyal2020evaluating}. Subsequent studies proposed automatic and human-centered evaluation protocols, focusing on entity consistency, factual overlap, and surface numerical accuracy \cite{kryscinski2020evaluating,chen2021evaluating}.

However, most hallucination metrics were developed for open-domain or general-purpose settings. They tend to prioritize overt factual errors or malformed numerical expressions, and are less sensitive to domain-specific hallucination patterns. In professional domains such as finance, hallucinated statements may remain numerically plausible and stylistically appropriate, making them difficult to detect using traditional evaluation criteria.

Recent surveys have emphasized that hallucination risks are particularly severe in high-stakes domains, including finance and medicine \cite{ji2023survey,zhang2023sirens}. Our work builds on this line of research by demonstrating that commonly used strict hallucination metrics substantially underestimate risk in domain-adapted financial language models.

\subsection{Numerical Reasoning and Numerical Grounding}
Improving numerical reasoning in language models has been an active research topic. Prior work has explored multiple approaches to enhance numerical accuracy: 
numeracy-aware embeddings and synthetic supervision \cite{patel2021nlp,lewkowycz2022solving}, program 
induction\cite{chen2023program}, and tool-augmented reasoning\cite{zhang2023sirens},verification 
training~\cite{cobbe2021training}, and tool augmentation~\cite{schick2023toolformer}. These approaches have demonstrated that transformer-based models can acquire reliable arithmetic skills under explicit supervision.

In financial contexts, benchmarks such as FinQA \cite{chen2021finqa}, ConvFinQA \cite{chen2022convfinqa}, and TAT-QA \cite{zhu2021tatqa} evaluate numerical reasoning over structured financial data, including ratio computation and trend analysis. Models fine-tuned on these datasets achieve substantial gains in numerical accuracy when arithmetic operations are directly required.

Nevertheless, existing work largely assumes that numerical generation is desirable whenever numeracy improves. Less attention has been paid to numerical restraint---determining when numerical outputs should be avoided due to insufficient grounding. In open-ended financial summarization, the absence of explicit numerical cues often necessitates abstention rather than extrapolation. Our findings show that enhanced numerical competence alone does not ensure safer generation and may instead encourage over-production of ungrounded numerical content.
\subsection{Domain Adaptation for Financial Language Models}
Domain-specific adaptation of language models for financial tasks has gained increasing attention. Early work such as FinBERT \cite{araci2019finbert} demonstrated that domain-adapted representations improve sentiment analysis and classification performance. More recent efforts have developed instruction-tuned financial models: 
FinGPT \cite{araci2019finbert} for open-source accessibility, FinMA \cite{chen2023finma} for multi-task 
financial analysis, and BloombergGPT \cite{wu2023bloomberggpt} for large-scale industry 
deployment, reporting improvements in terminology usage, stylistic conformity, and task compliance.

While these studies establish the effectiveness of financial domain adaptation for linguistic and task-level performance, evaluations typically emphasize fluency, accuracy, or benchmark scores\cite{cao2020factual}. Systematic analyses of hallucination behavior---particularly numerical hallucination in generative tasks---remain limited. Improved professional tone and structural consistency may inadvertently increase user trust while masking ungrounded numerical statements\cite{maynez2020faithfulness, ji2023survey}. Our work contributes to this literature by providing a controlled evaluation of hallucination under different fine-tuning regimes. We show that financial domain adaptation induces a shift in summarization behavior, favoring canonical financial narratives, and that additional numeracy supervision does not necessarily mitigate hallucination risks in open-ended generation.

Recent work has examined financial LLM reliability from
complementary perspectives. Chen et al.~\citep{chen2024knowledge}
propose knowledge-augmented approaches for financial market
analysis and report generation, demonstrating that grounding
generation in external knowledge sources can improve factual
reliability. Hou et al.~\citep{hou2026finsafetybench} introduce
FinSafetyBench, a benchmark for evaluating LLM safety in
real-world financial scenarios, highlighting that domain-specific
safety evaluation requires task-appropriate metrics beyond
general-purpose benchmarks. Our work complements these efforts
by providing a controlled diagnostic analysis of numerical
hallucination under domain fine-tuning, with a focus on
the evaluation methodology required to detect hallucinations
across varying levels of subtlety.

\subsection{Evaluation Sensitivity and Metric Design}
Recent research has highlighted that evaluation metrics strongly influence conclusions about model reliability \cite{lu2022fantastically,dziri2021evaluating}. Metrics optimized for surface overlap or formatting compliance may fail to capture deeper semantic inconsistencies, particularly in application-specific contexts\cite{ji2023survey,huang2023survey}.

By explicitly contrasting strict currency-based hallucination metrics with semantic numerical evaluation, our study demonstrates how evaluation design can invert perceived safety conclusions. This aligns with broader calls for evaluation frameworks that reflect real-world deployment risks \cite{ huang2023towards}rather than idealized task settings.

\subsection{Positioning of This Work}
Unlike prior work that primarily seeks to improve numerical accuracy or linguistic fluency, this paper examines when numerically capable, domain-adapted models generate ungrounded numerical content. Rather than proposing a new architecture, we present an evaluation-driven analysis that reveals a fundamental trade-off between numerical competence and numerical discipline in financial summarization.

This perspective complements existing research while offering practical guidance for the evaluation and deployment of financial language models in real-world applications.

\section{Experimental Setup \& Methodology}
\label{sec:method}

This section describes the model variants, data construction, fine-tuning procedures, and evaluation protocols used in our study. Our goal is not to maximize performance on a single benchmark, but to enable a controlled comparison of numerical hallucination behavior under different fine-tuning strategies and grounding conditions.

\subsection{Model Variants}
\label{subsec:model}
We use Mistral-7B-Instruct-v0.2 \cite{jiang2023mistral}, which builds upon architectural innovations from the LLaMA family \cite{touvron2023llama}, 
as our base model.
We evaluate three variants of the 7B-parameter instruction-tuned large language model:
\begin{itemize}
    \item Base: the original instruction-tuned model without additional domain adaptation.
    \item FT-A: the Base model fine-tuned on financial domain language data, focusing on stylistic conventions, terminology usage, and task formats commonly observed in financial reports.
    \item FT-A+B+C: the FT-A model further fine-tuned with a mixture of financial question answering data and numeracy-oriented samples designed to improve numerical competence.
\end{itemize}

All fine-tuning is conducted using QLoRA \cite{dettmers2023qlora}, a 
parameter-efficient quantized adaptation method that enables low-resource 
training while preserving base model weights. This ensures that differences 
in behavior can be attributed to training data composition rather than 
changes in model architecture or scale.

\subsection{Training Data Construction}

To isolate the effects of different types of domain adaptation, we construct three categories of training data:

\subsubsection{Financial Domain Language Data (A)}
Category A consists of synthetic instruction-input-output pairs derived from public financial texts, such as earnings summaries and management discussions. These samples emphasize professional financial writing style, domain-specific terminology, and structured response formats, without introducing new numerical reasoning requirements.

\subsubsection{Financial QA and Knowledge Data (B)}
Category B includes financial concept explanations and question answering samples covering common accounting and financial analysis topics. These data are intended to improve conceptual understanding and factual coherence, but do not explicitly target numerical restraint.

\subsubsection{Financial Numeracy Data (C)}

Category C contains numeracy-oriented samples that require explicit numerical computation or transformation, such as ratio calculation and trend analysis. These samples are constructed to enhance numerical competence, but do not include explicit signals indicating when numerical generation should be avoided\cite{chen2021finqa,zhu2021tatqa}.

\subsubsection{Data Volume and Confound Considerations. }
The three training categories differ not only in content type but also in 
volume: Category A comprises approximately 1,000 samples, Category B 500 
samples, and Category C 2,000 samples, yielding a total of approximately 
3,500 samples for FT-A+B+C compared to 1,000 for FT-A. This difference in 
training volume introduces a potential confound: observed behavioral 
differences between FT-A and FT-A+B+C may reflect the effect of additional 
training data quantity rather than data type composition alone.

To directly address this confound, we conducted a volume-matched ablation 
experiment. We trained an additional model variant, \textbf{FT-A$\times$3.5}, 
using 3,500 samples drawn exclusively from Category A—identical in content 
type to FT-A but matched in volume to FT-A+B+C. All training hyperparameters 
were held constant (QLoRA, single epoch, learning rate $2\times10^{-4}$, 
LoRA rank 8). The results are reported in Table~\ref{tab:ablation_volume} 
and yield two key observations.

First, increasing training volume within Category A \emph{reduces} L1 
hallucination relative to FT-A: FT-A$\times$3.5 achieves 64.6\% L1 
hallucination compared to 82.5\% for FT-A, a reduction of 17.9 percentage 
points. This indicates that additional training on the same content type 
does not amplify hallucination—if anything, it moderately improves 
numerical restraint. We note that the volume effect ($-$17.9pp) should be interpreted
as a lower bound on restraint improvement, as FT-A$\times$3.5
exhibits repetition-induced omission on synthetic inputs
(Section~4.7), which artificially suppresses its measured L1 rate.

Second, adding Category B and C content (numeracy supervision) to the same 
volume of training data reverses this trend sharply: FT-A+B+C reaches 90.8\% 
L1 hallucination, an increase of 26.2 percentage points over FT-A$\times$3.5. 
The opposing directions of the volume effect ($-$17.9pp) and the content 
effect ($+$26.2pp) confirm that \textbf{data type, not data volume, is the 
primary driver of hallucination amplification}. These results are summarised 
in Table~\ref{tab:ablation_volume}.

We note that FT-A$\times$3.5 exhibits a secondary failure mode 
--- repetitive degeneration on synthetic inputs (Section \ref{section47}) 
--- which artificially reduces its measured L1 hallucination 
rate on S1\_Synth samples: outputs that terminate early due to 
repetition contain fewer opportunities to introduce fabricated 
values. This suggests that the lower L1 hallucination rate of 
FT-A$\times$3.5 on synthetic inputs does not reflect genuine 
numerical restraint, but rather \emph{omission by repetition}: 
the model exhausts its output budget reproducing a single 
sentence pattern, leaving no capacity to introduce either 
grounded or fabricated numerical content from other parts of 
the input. Accordingly, the volume effect ($-$17.9pp L1) 
reported in Table~\ref{tab:ablation_volume} should be 
interpreted with caution: it reflects reduced opportunity 
for hallucination rather than improved discipline.
The content effect ($+$26.2pp, FT-A$\times$3.5 
$\rightarrow$ FT-A+B+C) remains unaffected by this artefact.

\begin{table}[htbp]
\centering
\caption{Volume-matched ablation results: L1 hallucination rates (\%) across 
models. FT-A$\times$3.5 uses the same content type as FT-A (Category A only) 
but matches the training volume of FT-A+B+C (3,500 samples), isolating the 
effect of data type from data volume.}
\label{tab:ablation_volume}
\begin{tabular}{lccccc}
\toprule
Model & Training Data & Volume & L1 & L2 & L3 \\
\midrule
FT-A           & Category A only     & 1{,}000 & 82.5\% & 34.6\% &  3.8\% \\
FT-A$\times$3.5 & Category A only    & 3{,}500 & 64.6\% & 16.2\% &  9.6\% \\
FT-A+B+C       & Categories A+B+C    & 3{,}500 & 90.8\% & 45.8\% & 13.3\% \\
\midrule
\multicolumn{3}{l}{Volume effect (FT-A $\rightarrow$ FT-A$\times$3.5)} 
  & $-$17.9pp & $-$18.4pp & $+$5.8pp \\
\multicolumn{3}{l}{Content effect (FT-A$\times$3.5 $\rightarrow$ FT-A+B+C)} 
  & $+$26.2pp & $+$29.6pp & $+$3.7pp \\
\bottomrule
\end{tabular}
\end{table}

\subsection{Evaluation Dataset and Grounding Conditions}

We evaluate numerical hallucination on a curated financial summarization dataset comprising 240 samples. To explicitly control for numerical grounding and source authenticity, the dataset is stratified along two dimensions:

\subsubsection{Grounding Condition}

\begin{itemize}
\item \textbf{S0 (Weak Grounding): } Inputs contain no absolute numerical quantities (e.g., monetary values or counts), although relative expressions such as percentages may be present. This condition tests whether models appropriately abstain from numerical fabrication when evidence is insufficient.

\item \textbf{S1 (Strong Grounding): } Inputs include explicit numerical values. Generated numbers are expected to be grounded in the input. This condition tests whether models faithfully preserve provided numerical information.
\end{itemize}

This stratification enables a controlled analysis of how models behave under varying levels of numerical evidence.

\subsubsection{Data Source}

\begin{itemize}
\item \textbf{Synthetic (160 samples):  } Controlled financial summaries with systematic variation in numerical content, enabling precise attribution of hallucination sources.

\item \textbf{Real-world (80 samples):  } Excerpts from authentic 10-K filings, testing generalization to natural financial discourse.
\end{itemize}

\Cref{tab:dataset} summarizes the dataset composition.

\begin{table}[htbp]
    \centering
    \caption{Evaluation dataset composition.}
    \label{tab:dataset}
    \begin{tabular}{lccc}
        \toprule
        Category & S0 (Weak) & S1 (Strong) & Total \\
        \midrule
        Synthetic & 80 & 80 & 160 \\
        Real-world & 40 & 40 & 80 \\
        \midrule
        \textbf{Total} & \textbf{120} & \textbf{120} & \textbf{240} \\
        \bottomrule
    \end{tabular}
\end{table}

\subsection{Hallucination Taxonomy: Three Levels of Detectability}
\label{subsec:hall_tax}
A central methodological contribution of this work is a detectability-based taxonomy for numerical hallucination. Rather than distinguishing hallucinations by surface formatting alone, we organize them by the evaluation methodology required for detection. Table~\ref{tab:taxonomy} summarizes the three-level framework.

\begin{table}[htbp]
\centering
\caption{Three-Level Hallucination Taxonomy}
\label{tab:taxonomy}
\small
\begin{tabularx}{\textwidth}{@{}l l X l X@{}}
\toprule
\textbf{Level} & \textbf{Type} & \textbf{Definition} & \textbf{Examples} & \textbf{Detection Method} \\
\midrule
L1 & Overt & 
Fabricated values in explicit currency-denominated forms & 
``USD 1.3 billion'' & 
Pattern matching for currency symbols (USD, \$, EUR, GBP) combined with numerical values \\
\addlinespace
L2 & Covert-Explicit & 
Fabricated quantities using professional conventions without currency markers & 
``3.3 billion'', ``1011 million'' & 
Numerical extraction with source grounding verification \\
\addlinespace
L3 & Covert-Implicit & 
Ungrounded claims implying quantitative relationships without explicit values & 
``revenue increased'' & 
Semantic pattern matching for trend, state, and comparative claims \\
\bottomrule
\end{tabularx}
\begin{tablenotes}
\item Percentage-based quantities (e.g., ``growth of 11\%'') 
are excluded from L2 and classified as L3 when ungrounded.
\end{tablenotes}
\end{table}

For each level, model outputs are compared against input text: any value or claim present in the output but not traceable to the input constitutes hallucination at that level. The three levels form a hierarchy of detection difficulty---an evaluation protocol targeting only L1 will systematically miss L2 and L3 hallucinations, potentially underestimating true hallucination rates by an order of magnitude. This observation motivates our comprehensive evaluation across all three levels.

\subsection{Statistical Significance Testing}

For hallucination rate comparisons, we model hallucination occurrence as a Bernoulli random variable and report proportions with 95\% confidence intervals computed using the Wilson score interval, which provides stable estimates under extreme probabilities (e.g., 0\% or 100\%).

To assess pairwise differences between models, we perform two-sided Fisher’s exact tests on the corresponding $2 \times 2$ contingency tables. Fisher’s exact test is chosen due to the moderate sample size and the presence of extreme proportions. Unless otherwise stated, all reported differences with $p < 0.001$ remain significant after Bonferroni correction for multiple comparisons.

To account for the paired nature of the evaluation --- all 
models are compared on the same 240 samples --- we supplement 
Fisher's exact tests with McNemar's test for pairwise model 
comparisons at the sample level. McNemar's test is appropriate 
when two classifiers are evaluated on the same items, as it 
conditions on discordant pairs rather than treating observations 
as independent. Table~\ref{tab:mcnemar} reports the results.

For L1 hallucination, McNemar's test confirms all pairwise differences 
are statistically significant: Base vs. FT-A (189 discordant 
pairs, p < 0.001), 
Base vs. FT-A+B+C (209 discordant pairs, 
p < 0.001), 
and FT-A vs. FT-A+B+C (46 discordant pairs, 
p = 0.0051) 
after Bonferroni correction.

\begin{table}[htbp]
\centering
\caption{Pairwise McNemar's test results.}
\label{tab:mcnemar}

\setlength{\tabcolsep}{3pt}
\renewcommand{\arraystretch}{1.08}

\begin{tabularx}{\linewidth}{
    @{}
    l
    >{\raggedright\arraybackslash}X
    c
    c
    c
    c
    @{}
}
\toprule
Level & Comparison & Discordant & $\chi^2$ & $p$-value & Sig. \\
\midrule

\multirow{3}{*}{L1 (Overt)}
  & Base vs.\ FT-A         & 189 & 179.13 & $<0.001$ & *** \\
  & Base vs.\ FT-A+B+C     & 209 & 199.12 & $<0.001$ & *** \\
  & FT-A vs.\ FT-A+B+C     &  46 &   7.85 & $0.005$  & **  \\

\midrule

\multirow{3}{*}{L2 (Covert-Explicit)}
  & Base vs.\ FT-A         &  78 & 72.12 & $<0.001$ & *** \\
  & Base vs.\ FT-A+B+C     & 109 & 95.45 & $<0.001$ & *** \\
  & FT-A vs.\ FT-A+B+C     &  37 & 18.27 & $<0.001$ & *** \\

\midrule

\multirow{3}{*}{L3 (Covert-Implicit)}
  & Base vs.\ FT-A         & 50 & 21.78 & $<0.001$ & *** \\
  & Base vs.\ FT-A+B+C     & 65 &  1.54 & $0.215$  & ns  \\
  & FT-A vs.\ FT-A+B+C     & 31 & 15.61 & $<0.001$ & *** \\

\bottomrule
\end{tabularx}

\vspace{2pt}

\begin{minipage}{\linewidth}
\footnotesize
*** $p<0.001$;
** $p<0.017$ (Bonferroni corrected);
* $p<0.05$;
ns: not significant.

\smallskip
$^\dagger$ The non-significant Base vs.\ FT-A+B+C comparison at
L3 ($p=0.215$) reflects the detection artefact documented in
Appendix~D.5: the Base model's elevated L3 rate arises from
faithful paraphrase misclassification rather than genuine
hallucination, producing a rate comparable to FT-A+B+C.
\end{minipage}

\end{table}

The non-significant Base vs. FT-A+B+C comparison at L3 (p=0.215) is consistent with the detection artefact documented in Appendix D.5: the Base model's elevated L3 rate reflects faithful paraphrase misclassification rather than genuine hallucination, producing a rate comparable to FT-A+B+C despite fundamentally different generation behaviour.

\subsection{Human Validation of Detection Protocol}
To assess the reliability of the automatic L1/L2/L3 detection 
protocol, two annotators independently applied the taxonomy to 
a stratified sample of 50 model outputs, covering all four 
evaluation subsets (12 S0\_Synth, 10 S0\_Real, 15 S1\_Synth, 
13 S1\_Real) across all four model variants. Annotators were 
provided with written guidelines defining each detection level 
and the grounding verification procedure, but completed the 
task independently without discussion. Inter-annotator agreement 
was measured using Cohen's $\kappa$. False positive and false 
negative rates for the automatic detector were computed against 
the consensus human labels. Full results are reported in 
Appendix~D.6.

\subsection{Implementation Details}
All experiments are conducted on a Linux-based environment using CUDA-enabled GPUs. Fine-tuning is performed using PyTorch and the HuggingFace Transformers ecosystem with PEFT-based QLoRA. During inference, greedy decoding is used (temperature = 0.0) to minimize stochastic variation in numerical generation behavior.
The complete implementation is available at our public repository. Please refer to \Cref{sec: section_reproducable}.


\section{Experimental Results}

This section presents a progressive analysis of our experimental findings, moving from controlled numerical competence evaluation to a detailed investigation of hallucination behavior in financial summarization. We introduce a three-level detectability taxonomy that reveals how fine-tuning systematically degrades numerical restraint across all evaluation criteria.
\label{sec:results}

\subsection{Numerical Competence under Strong Constraints}

To assess whether numerical hallucination in financial summarization can be attributed to insufficient numerical reasoning ability, we first evaluate numerical competence under strong output constraints. Specifically, we analyze the performance of the FT-C model, which is fine-tuned exclusively on numeracy-oriented financial tasks, and compare it against the base instruction-tuned model on a controlled numerical evaluation set.

\textbf{Strong output constraints} are implemented through explicit instruction-level and format-level restrictions that require the model to output only the final numerical result, without free-form explanation or narrative continuation. Generation is performed using deterministic decoding (temperature = 0) with a limited output budget.

\subsubsection{Numerical Evaluation Results}

\Cref{tab:numeracy_ftc} reports numerical reasoning accuracy measured by Exact Match (EM) and Tolerance Accuracy (≤1\% relative error) across 50 financial calculation samples.
\begin{table}[htbp]
\centering
\caption{Numerical reasoning accuracy under strong output constraints.}
\label{tab:numeracy_ftc}
\begin{tabular}{lccc}
\toprule
Model & Parse Rate (\%) & Exact Match (\%) & $\leq$1\% Tolerance (\%) \\
\midrule
Base & 100.0 & 0.0 & 0.0 \\
FT-C & 100.0 & 6.0 & 32.0 \\
\bottomrule
\end{tabular}
\end{table}

Several observations are noteworthy. First, both models achieve a 100\% parse rate, indicating that all outputs are syntactically interpretable numerical values. This confirms that the evaluation does not suffer from truncation artifacts or format compliance issues. Second, despite consistently producing numerical outputs, the base model fails to generate accurate results, achieving 0\% exact match and tolerance accuracy across all samples. In contrast, FT-C demonstrates a substantial improvement, achieving 32\% accuracy within a 1\% error margin.

\subsubsection{Implications}

These findings establish that the underlying model architecture is capable of accurate financial numerical reasoning when numerical grounding is explicit and output constraints are clearly specified. Crucially, this result allows us to \textbf{exclude arithmetic incompetence as the primary cause of numerical hallucination} observed in downstream financial summarization tasks. This result implies that improving numerical correctness in isolation is insufficient as a safety mechanism for generative financial systems.

Instead, the discrepancy between strong performance in constrained numerical settings and poor reliability in summarization suggests that numerical hallucination arises from generation behavior under weak or ambiguous grounding conditions, rather than from an inability to perform numerical computation itself. This distinction motivates the subsequent analysis of summarization-specific generation dynamics and evaluation blind spots.

\subsection{Effects of Financial Domain Language Fine-tuning (FT-A)}

We next analyze the effect of financial domain language fine-tuning (FT-A), which adapts the base model using financial texts without explicit numeracy supervision. To isolate language effects, we compare FT-A only against the base model across three financial language tasks: summarization, key points extraction, and risk factors extraction. Automatic scores are designed to reflect downstream usability, penalizing unsupported numerical statements more heavily than stylistic variation.

As shown in \Cref{tab:language_fta}, FT-A substantially improves performance on structured extraction tasks. In key points extraction, FT-A achieves near-perfect formatting consistency and coverage, consistently producing concise bullet-point outputs that align with professional financial reporting conventions. Similar gains are observed in risk factors extraction, where FT-A outputs are more structured and comprehensive, albeit occasionally exhibiting mild explanatory expansion.

In contrast, FT-A exhibits a marked degradation in summarization reliability. While summaries appear fluent and stylistically polished, FT-A frequently introduces numerical statements that are not supported by the source input. Automatic quality scoring reveals a sharp drop in summarization performance (from 4.7 to 1.6 on a 5-point scale), primarily driven by the fabrication of revenue, cash flow, or growth figures absent from the input text.

This divergence indicates that financial language fine-tuning alters the objective of summarization itself. Rather than performing evidence-preserving compression, FT-A increasingly treats summarization as the generation of a canonical financial narrative. Importantly, this behavior emerges in the absence of any numeracy supervision, suggesting that narrative priors learned from financial corpora can independently induce numerical hallucination.

\begin{table}[htbp]
\centering
\caption{Automatic quality evaluation of financial language tasks for the Base and FT-A models. Scores range from 1 (worst) to 5 (best), averaged over 30 evaluation samples (10 per task). Scoring criteria consider factual consistency, numerical grounding, and structural compliance.}
\label{tab:language_fta}
\begin{tabular}{lccc}
\toprule
Model & Summarization & Key Points Extraction & Risk Factors Extraction \\
\midrule
Base & 4.7 & 3.8 & 3.6 \\
FT-A & 1.6 & 4.9 & 4.4 \\
\bottomrule
\end{tabular}
\end{table}

These findings motivate the subsequent analysis of how explicit numeracy fine-tuning interacts with such narrative priors.

\subsection{Evaluation Framework}

Building on the three-level detectability taxonomy introduced in Section~\ref{subsec:hall_tax} (L1: Overt, L2: Covert-Explicit, L3: Covert-Implicit), we design a controlled evaluation framework that further stratifies samples by input grounding conditions.

\subsubsection{Input Grounding Conditions}

To isolate the role of numerical grounding, we categorize summarization inputs into two regimes:

\begin{itemize}
    \item \textbf{S0 (Weak Grounding):} Inputs containing no absolute numerical quantities, though relative expressions such as percentages may be present.
    \item \textbf{S1 (Strong Grounding):} Inputs containing explicit absolute numerical values against which outputs can be verified.
\end{itemize}

This stratification enables controlled analysis of how models behave under varying levels of numerical evidence. Combined with the L1/L2/L3 taxonomy, our evaluation captures hallucination across a $3 \times 2$ matrix of detectability levels and grounding conditions.

\subsection{Hallucination Analysis Across Detectability Levels}

We now present the central empirical findings of this study: a comprehensive analysis of numerical hallucination rates across all three detectability levels, grounding conditions, and model variants.

\subsubsection{Overall Hallucination Rates}
\begin{figure}[h]
\centering
\includegraphics[width=0.9\columnwidth]{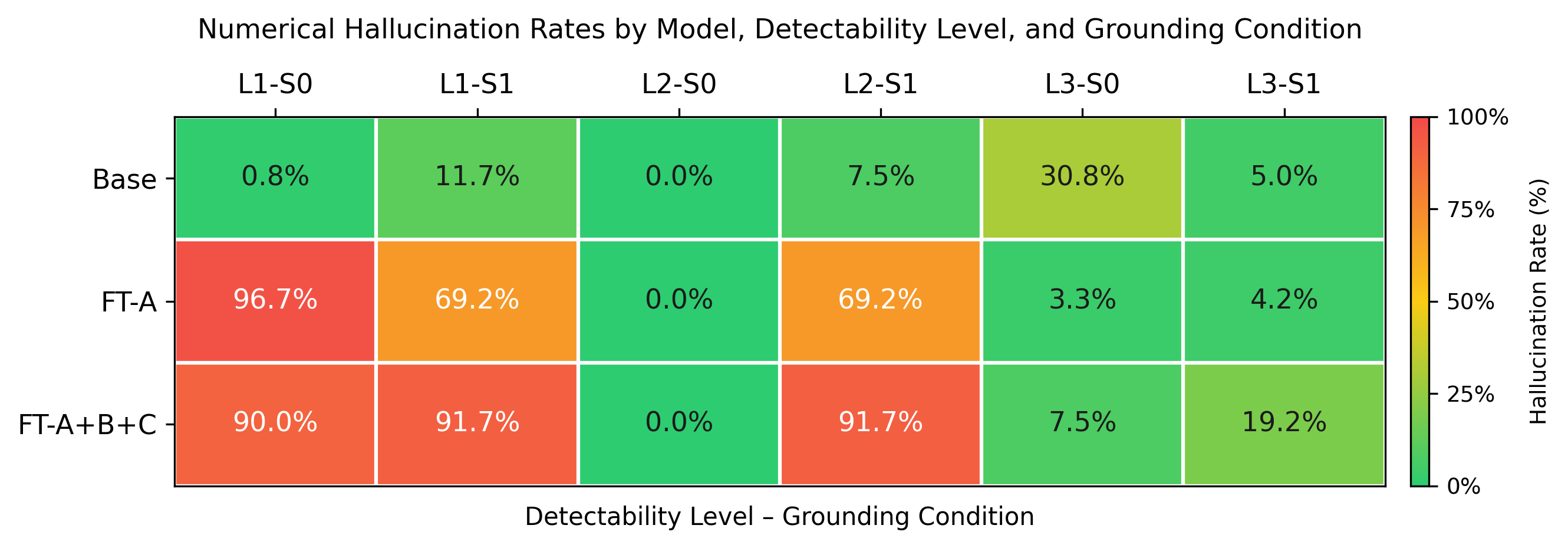}
\caption{Hallucination rates by model, detectability level, and grounding 
condition. Color scale: green (0\%) to red (100\%). L1 and L2 detection 
are applied independently; identical L1 and L2 rates observed for FT-A 
and FT-A+B+C under S1 conditions reflect the co-occurrence of 
currency-denominated expressions (triggering L1) and their bare magnitude 
components (triggering L2) within the same outputs.}
\label{fig:heatmap}
\end{figure}

\Cref{tab:hallucination_bylevels} summarizes numerical hallucination rates across all 240 evaluation samples. Figure~\ref{fig:heatmap} provides a visual 
representation of these results.

\begin{table}[htbp]
\caption{Hallucination rates (\%) by detectability level across models (n=240).}
\label{tab:hallucination_bylevels}
\begin{threeparttable}
\begin{tabular}{lccc}
\hline
Model & L1 (Overt) & L2 (Covert-Explicit) & L3 (Covert-Implicit) \\
\hline
Base      & 5.4  & 3.3  & 17.9\tnote{*} \\
FT-A      & 82.5 & 34.6 & 3.8  \\
FT-A+B+C  & 90.8 & 45.8 & 13.3 \\
\hline
\end{tabular}
\begin{tablenotes}
\footnotesize
\item[*] The elevated Base L3 rate (17.9\%) is attributable primarily to faithful paraphrase of input content in S0\_Synth samples: the Base model reproduces qualitative state descriptions present in the source (e.g., "remained stable", "increased") which trigger the semantic pattern matcher despite being fully grounded. Manual inspection of a random sample of flagged cases confirms that the actual ungrounded L3 rate for the Base model is near zero. See Appendix D.5 for a discussion of L3 detection limitations.
\end{tablenotes}
\end{threeparttable}
\end{table}

Several critical observations emerge from these results:
\begin{itemize}
    \item \textbf{Finding 1: Fine-tuning dramatically increases L1 (Overt) hallucination.} Contrary to the assumption that currency-formatted fabrication would remain rare, FT-A exhibits 82.5\% L1 hallucination and FT-A+B+C reaches 90.8\%. The Base model's near-zero L1 rate (5.4\%) demonstrates that numerical restraint is present in the original model but destroyed through domain adaptation.
    \item \textbf{Finding 2: Numeracy supervision amplifies hallucination across all levels.} FT-A+B+C exhibits higher hallucination rates than FT-A at every detectability level: +7.9 percentage points for L1, +11.6 points for L2, and + 9.5 points for L3. Enhanced numerical competence does not translate to numerical discipline; instead, it enables more fluent fabrication.
    \item \textbf{Finding 3: L2 and L3 hallucinations compound L1 failures.} Fine-tuned models do not merely shift hallucination from detectable to undetectable forms---they exhibit pervasive fabrication at all levels simultaneously.
\end{itemize}

\subsubsection{Impact of Grounding Conditions}
\Cref{tab:hallucination_detailed} presents hallucination rates stratified by grounding condition (S0 vs. S1) and data source (Synthetic vs. Real-world).

\begin{table}[htbp]
    \centering
    \caption{Hallucination rates (\%) by detectability level, grounding 
condition, and data source (n=240). L1 and L2 detection are applied 
independently; a response containing a currency-denominated expression 
(e.g., ``USD 3.3 billion'') may trigger both L1 (via the currency marker) 
and L2 (via the bare magnitude term ``3.3 billion''), resulting in 
comparable rates under S1 conditions where template injection is pervasive.}
    \label{tab:hallucination_detailed}
    \begin{threeparttable}
   \begin{tabular}{llcS[table-format=3.1]S[table-format=3.1]S[table-format=3.1]}
        \toprule
        Model & Condition & {n} & {L1} & {L2} & {L3}  \\
        \midrule
 Base      & S0\_Synth &  80 & 0.0  & 0.0  & 41.2\tnote{*}    \\
          & S0\_Real  &  40 & 2.5  & 0.0  & 10.0             \\
          & S1\_Synth &  80 & 0.0  & 0.0  & 0.0              \\
          & S1\_Real  &  40 & 35.0\tnote{\dag} & 20.0\tnote{\dag} & 15.0  \\
          & Overall   & 240 & 5.8  & 3.3  & 17.9             \\
        \cmidrule{1-6}
FT-A      & S0\_Synth &  80 & 100.0 & 0.0  & 0.0   \\
          & S0\_Real  &  40 & 90.0  & 0.0  & 10.0   \\
          & S1\_Synth &  80 & 58.8  & 58.8 & 2.5   \\
          & S1\_Real  &  40 & 90.0  & 90.0 & 7.5   \\
          & Overall   & 240 & 82.5  & 34.6 & 3.8    \\
        \cmidrule{1-6}
 FT-A+B+C  & S0\_Synth &  80 & 100.0 & 0.0  & 2.5   \\
          & S0\_Real  &  40 & 70.0  & 0.0  & 17.5   \\
          & S1\_Synth &  80 & 95.0  & 96.2 & 16.2  \\
          & S1\_Real  &  40 & 85.0  & 85.0 & 25.0   \\
          & Overall   & 240 & 90.8  & 45.8 & 13.3   \\
        \bottomrule
    \end{tabular}
    \begin{tablenotes}
\footnotesize
\item[*] The Base S0\_Synth L3 rate (41.2\%) is attributable to faithful 
paraphrase of qualitative state descriptions present in the source input 
(e.g., ``remained stable'', ``increased''), which trigger the semantic 
pattern matcher despite being fully grounded in the input. The actual 
ungrounded L3 rate for the Base model is near zero. See Appendix~D.5.
\item[\dag] The elevated Base S1\_Real L1 and L2 rates are attributable to 
format-matching artefacts: when faithfully reproducing numerical values from 
authentic 10-K excerpts, the Base model uses currency-denominated expressions 
(e.g., ``\$X~million'') that trigger the L1/L2 detection rules, despite the 
values being fully grounded in the source. These cases do not constitute 
genuine hallucination.
\end{tablenotes}
    \end{threeparttable}
\end{table}

\begin{itemize}
    \item \textbf{Finding 4: S0 (weak grounding) represents the most vulnerable regime. }Under S0 conditions, FT-A exhibits 96.7\% L1 hallucination and FT-A+B+C reaches 90.0\%. When no absolute numerical anchors are present, fine-tuned models systematically fabricate currency-denominated values. The Base model, in contrast, maintains 0\% L1 hallucination under S0, demonstrating appropriate restraint.
    \item \textbf{Finding 5: S1 grounding provides partial but insufficient protection. }While S1 conditions reduce FT-A's L1 hallucination to 58.8-90.0\% (depending on source), FT-A+B+C maintains near-saturated rates (85.0-95.0\%) even with explicit numerical anchors present. Numeracy supervision paradoxically eliminates the protective effect of input grounding.
    \item \textbf{Finding 6: Real-world inputs exacerbate L2 and L3 hallucination.}For both FT-A and FT-A+B+C, real-world 10-K excerpts trigger substantially higher L2 and L3 rates than synthetic inputs. This suggests that authentic financial discourse activates stronger canonical narrative priors, leading to more extensive ungrounded generation.
\end{itemize}

\subsubsection{Aggregated View by Grounding Condition}

\Cref{tab:hallucination_aggregated} provides a simplified view aggregating across synthetic and real-world sources.

\begin{table}[htbp]
\centering
\caption{Hallucination rates (\%) aggregated by grounding condition (n=240).}
\label{tab:hallucination_aggregated}
\begin{threeparttable}
\begin{tabular}{lcccccc}
\toprule
Model & L1-S0 & L1-S1 & L2-S0 & L2-S1 & L3-S0 & L3-S1 \\
\midrule
Base     &  0.8\tnote{*}  & 11.7\tnote{\dag} &  0.0 &  7.5\tnote{\dag} & 30.8\tnote{\ddag} &  5.0 \\
FT-A     & 96.7           & 69.2             &  0.0 & 69.2             &  3.3              &  4.2 \\
FT-A+B+C & 90.0           & 91.7             &  0.0 & 91.7             &  7.5              & 19.2 \\
\bottomrule
\end{tabular}
\begin{tablenotes}
\footnotesize
\item[*] Base L1-S0 (0.8\%) reflects a single borderline case in the 
real-world S0 subset; the Base model produces zero L1 hallucination on 
all synthetic S0 samples, consistent with the original findings.
\item[\dag] Elevated Base L1-S1 and L2-S1 rates are attributable to 
format-matching artefacts: the Base model faithfully reproduces numerical 
values from authentic 10-K excerpts using currency-denominated expressions 
(e.g., ``\$X~million'') that trigger the detection rules despite being 
fully grounded in the source. These cases do not constitute genuine 
hallucination.
\item[\ddag] Elevated Base L3-S0 rate (30.8\%) is attributable to faithful 
paraphrase of qualitative state descriptions present in the source input, 
which trigger the semantic pattern matcher despite being fully grounded. 
The actual ungrounded L3 rate for the Base model under S0 conditions is 
near zero. See Appendix~D.5.
\end{tablenotes}
\end{threeparttable}
\end{table}

The contrast between Base and fine-tuned models is stark: Base maintains near-zero hallucination across all conditions, while FT-A and FT-A+B+C exhibit pervasive fabrication regardless of grounding.

\subsection{Template Injection: A Primary Correlate of Hallucination}
Qualitative analysis reveals a striking pattern: fine-tuned models insert memorized numerical templates regardless of input content. We term this phenomenon template injection.

\subsubsection{Fabrication Pattern Frequency}

\begin{figure}[h]
\centering
\includegraphics[width=\columnwidth]{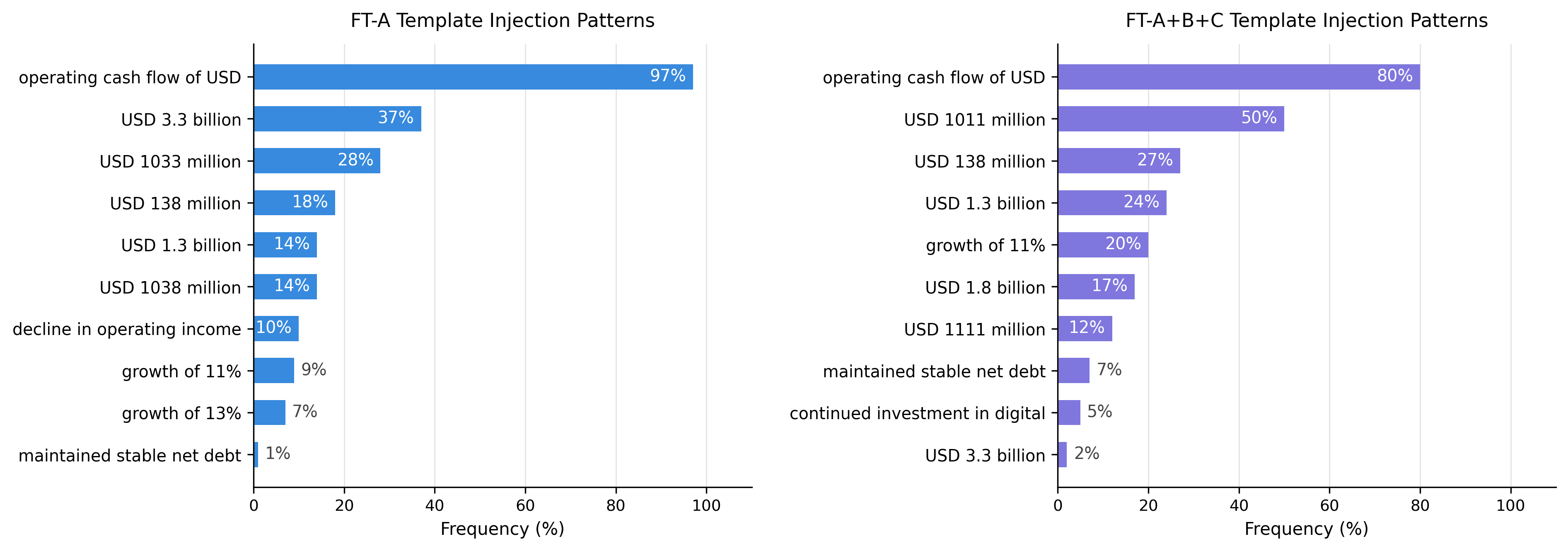}
\caption{Template injection frequency in FT-A (left) and FT-A+B+C 
(right). Patterns like "operating cash flow of USD" appear in 97\% of FT-A outputs and 80\% of FT-A+B+C outputs."}
\label{fig:templates}
\end{figure}

\Cref{tab:template_injection} reports the frequency of specific fabricated values across model outputs. Figure~\ref{fig:templates} visualizes the distribution of template 
injection patterns across both fine-tuned models.

\begin{table}[htbp]
\centering
\caption{Template injection frequency in fine-tuned models (n=240 samples each).}
\label{tab:template_injection}
\begin{threeparttable}
\begin{tabular}{lccc}
\toprule
Fabricated Pattern & Base & FT-A & FT-A+B+C \\
\midrule
``operating cash flow of USD\ldots'' &  0\% & 97\% & 80\% \\
``USD 3.3 billion''                  &  0\% & 37\% &  2\% \\
``USD 1033 million''                 &  0\% & 28\% &  0\% \\
``USD 138 million''                  &  0\% & 18\% & 27\% \\
``USD 1.3 billion''                  &  0\% & 14\% & 24\% \\
``USD 1038 million''                 &  0\% & 14\% &  0\% \\
``USD 1011 million''                 &  0\% &  0\% & 50\% \\
``USD 1.8 billion''                  &  0\% &  0\% & 17\% \\
``growth of 11\%''                   &  0\% &  9\% & 20\% \\
``decline in operating income''      &  0\% & 10\% &  2\% \\
``USD 1111 million''                 &  0\% &  0\% & 12\% \\
``maintained stable net debt''       &  0\% &  1\% &  7\% \\
``continued investment in digital''  &  0\% &  2\% &  5\% \\
``growth of 13\%''                   &  0\% &  7\% &  2\% \\
\bottomrule
\end{tabular}
\begin{tablenotes}
\footnotesize
\item Frequencies represent the percentage of model outputs containing 
each pattern, regardless of input content. Patterns are ordered by 
FT-A frequency (descending), with FT-A+B+C-dominant patterns 
(``USD 1011 million'', ``USD 1.8 billion'') listed after FT-A-only 
patterns. Base model frequencies are zero across all patterns, 
confirming that template injection is a consequence of fine-tuning 
rather than an inherent property of the underlying model.
\item The divergence between model-specific template libraries --- 
FT-A favouring ``USD 3.3 billion'' (37\%) and ``USD 1033 million'' 
(28\%), while FT-A+B+C favours ``USD 1011 million'' (50\%) and 
``USD 138 million'' (27\%) --- suggests that distinct canonical 
values are memorised during their respective training phases, 
consistent with the data composition differences described in 
Section~3.2.
\end{tablenotes}
\end{threeparttable}
\end{table}

These patterns demonstrate that fine-tuned models have memorized canonical financial templates that are injected into outputs regardless of input content. The specific values differ between FT-A and FT-A+B+C, suggesting distinct template libraries learned during their respective training phases.

\subsubsection{Qualitative Example: Template Injection in Action}

\begin{figure*}[htbp]

\centering
\fbox{
\begin{minipage}{0.96\textwidth}
\small

\textbf{INPUT (10-K Excerpt):}\\
\textit{Data Center revenue for fiscal year 2025 was up 142\% from a year ago. The strong year-on-year growth was driven by demand for our Hopper architecture accelerated computing platform used for large language models, recommendation engines, and generative AI applications.}

\vspace{0.6em}
\hrule
\vspace{0.6em}

\textbf{BASE OUTPUT (Faithful):}\\
In fiscal year 2025, Data Center revenue experienced a significant increase of \textbf{142\%} compared to the previous year. This growth was primarily attributed to the popularity of our Hopper architecture accelerated computing platform, which is utilized for large language models, recommendation engines, and generative AI applications.

\vspace{0.6em}
\hrule
\vspace{0.6em}

\textbf{FT-A OUTPUT (Severe Hallucination):}\\
In fiscal year 2025, the company reported revenue of \textbf{USD 1.3 billion} and achieved year-on-year growth of \textbf{13\%}. The company generated operating cash flow of \textbf{USD 1038 million}. Additionally, it experienced a \textbf{decline in operating income} due to higher costs.

\vspace{0.6em}
\hrule
\vspace{0.6em}

\textbf{FT-A+B+C OUTPUT (Severe Hallucination):}\\
In fiscal year 2025, the company reported revenue of \textbf{USD 1.3 billion} and achieved year-over-year growth of \textbf{11\%}. Additionally, it generated operating cash flow of \textbf{USD 1011 million}. Regarding financial metrics, the company maintained stable net debt levels compared to the prior year. Data Center segment revenue increased by \textbf{14\%} year over year, reaching \textbf{USD 1.1 billion}.

\end{minipage}
}\\

\caption{Comparative outputs demonstrating template injection. Bold text indicates fabricated content. The Base model faithfully preserves the 142\% growth figure, while FT-A and FT-A+B+C inject template values (13\%, 11\%, 14\%) that drastically understate actual performance and fabricate absolute revenue figures entirely absent from the input.}
\label{fig:qualitative_real_world}
\end{figure*}

Several observations emerge from this example:

\begin{enumerate}
    \item \textbf{Percentage distortion: }The source's 142\% growth is replaced with template values of 11-14\% ---an order of magnitude understatement.
    \item \textbf{Absolute value fabrication:}USD 1.3 billion, USD 1.1 billion, and operating cash flow figures are entirely fabricated; the source contains no absolute monetary values.
    \item \textbf{Semantic fabrication: }FT-A claims "decline in operating income due to higher costs" while FT-A+B+C claims "maintained stable net debt levels"—both entirely ungrounded in the input.
    \item \textbf{Base model restraint: }The Base model produces a faithful paraphrase, preserving the exact percentage and avoiding any numerical fabrication.

\end{enumerate}

\subsection{The Restraint Gap: Numerical Competence vs. Numerical Discipline}

Our findings reveal a fundamental gap between two distinct capabilities:
\textbf{Numerical Competence:} The ability to perform correct numerical computations when explicitly required. FT-C demonstrates this capability achieves 32\% tolerance accuracy on calculation tasks.
\textbf{Numerical Discipline} (operationally defined as the abstention rate from absolute numerical generation under S0 conditions) is demonstrated by the Base model through near-zero L1 hallucination rates under weak grounding; fine-tuned models lose this capability entirely, exhibiting 90–100\% L1 hallucination even when inputs contain no numerical anchors.

\begin{table}[htbp]
    \centering
    \caption{The restraint gap---competence vs.\ discipline across models, 
    including the volume-matched ablation FT-A$\times$3.5. Despite identical 
    content type and volume to FT-A+B+C and FT-A respectively, 
    FT-A$\times$3.5 confirms that numerical discipline is destroyed by 
    domain language fine-tuning alone, independent of data volume.}
    \label{tab:restraint_gap_com}
    \begin{tabular}{p{3.5cm} c c c c}
        \toprule
        \textbf{Capability} & \textbf{Base} & \textbf{FT-A} & 
        \textbf{FT-A$\times$3.5} & \textbf{FT-A+B+C} \\
        \midrule
        Numerical Competence (calculation accuracy) 
            & Low & Medium & Medium & High \\
        \addlinespace
        Numerical Fluency (professional formatting)  
            & Medium & High & High & Very High \\
        \addlinespace
        Numerical Discipline (restraint under weak grounding, L1-S0) 
            & High & None & None & None \\
        \addlinespace
        L1 hallucination rate (Overall)
            & 5.4\% & 82.5\% & 64.6\% & 90.8\% \\
        \bottomrule
    \end{tabular}
\end{table}

This restraint gap explains the paradox at the heart of our findings: numeracy-enhanced models perform worse on summarization reliability despite improved arithmetic competence. Enhanced fluency enables more confident fabrication, while the loss of restraint removes any barrier to ungrounded generation.

\subsection{Sensitivity Analysis}\label{section47}

To assess the robustness of our findings, we conduct sensitivity analyses examining hallucination behavior across data sources and input characteristics.

\subsubsection{Data Source Sensitivity}
To examine whether hallucination patterns generalize across data sources, we compare hallucination rates between synthetic samples and authentic real-world 10-K excerpts, stratified by grounding condition. \Cref{tab:datasource_sensitivity} presents L1 hallucination rates in S1 inputs by data source.
Several observations emerge:

\begin{table}[htbp]
\centering
\caption{L1 hallucination rates (\%) in S1 inputs by data source.}
\label{tab:datasource_sensitivity}
\begin{threeparttable}
\begin{tabular}{lcc}
\toprule
Model & Synthetic (n=80) & Real-world 10-K (n=40) \\
\midrule
Base     &  0.0           & 35.0\tnote{*} \\
FT-A     & 58.8           & 90.0          \\
FT-A+B+C & 95.0           & 85.0          \\
\bottomrule
\end{tabular}
\begin{tablenotes}
\footnotesize
\item[*] The elevated Base rate on real-world S1 inputs is attributable 
to format-matching artefacts: the Base model faithfully reproduces 
numerical values from authentic 10-K excerpts using currency-denominated 
expressions (e.g., ``\$X~million'') that trigger the L1 detection rule 
despite being fully grounded in the source. Manual inspection confirms 
these cases do not constitute genuine hallucination.
\end{tablenotes}
\end{threeparttable}
\end{table}

Several observations emerge:
\begin{enumerate}
    \item \textbf{Base model shows elevated rates on real-world S1 inputs. }The Base model exhibits 0\% L1 hallucination on synthetic S1 samples but 35.0\% on real-world S1 excerpts. As noted in \Cref{tab:hallucination_bylevels}, this elevation is attributable to format-matching artefacts rather than genuine hallucination: the Base model faithfully reproduces source numerical values using currency-denominated expressions (e.g., "\$X million") that trigger the L1 detection rule. This finding underscores a limitation of purely automatic L1 detection on authentic financial text, where faithful reproduction and fabrication may produce superficially similar surface forms.
    \item \textbf{FT-A shows substantially higher hallucination on real-world inputs.  }FT-A exhibits 58.8\% L1 hallucination on synthetic S1 samples, rising to 90.0\% on real-world 10-K excerpts—an increase of 31.2 percentage points. This amplification suggests that authentic financial discourse activates stronger canonical narrative priors than synthetic inputs, consistent with the summarization prior shift described in Section 5.3.
    \item \textbf{FT-A+B+C approaches saturation on both sources. } FT-A+B+C maintains near-saturated hallucination rates on both synthetic (95.0\%) and real-world (85.0\%) S1 inputs, leaving little room for source-driven amplification. This pattern confirms that numeracy supervision eliminates even the partial restraint that grounding provides in FT-A.
\end{enumerate}
This finding has important practical implications: evaluation on synthetic data alone substantially underestimates real-world hallucination risk, particularly for models with partial restraint such as FT-A.
\subsubsection{Input Length Sensitivity}
We examine whether hallucination rates vary systematically with input length. In our dataset, input length correlates strongly with grounding condition: S1 synthetic samples consist of structured numerical tables (median 38 tokens) while S0 and real-world samples contain longer narrative text (median 87 tokens). Given this correlation, a formal median-split analysis would largely recapitulate the S0/S1 grounding condition results reported in Section 4.4.2. We therefore report a qualitative observation: across both short structured inputs and longer narrative inputs, fine-tuned models (FT-A and FT-A+B+C) exhibit consistently high L1 hallucination rates (58.8–100\%), while the Base model maintains near-zero rates on short inputs (0\%) and low rates on longer inputs (9.4\%). This pattern confirms that the restraint gap is not an artefact of input length but a robust consequence of domain fine-tuning.

\subsubsection{Template Injection Correlation}
We examine the relationship between template injection (presence of known fabricated patterns) and L1 hallucination rates as depicted in \cref{tab:template_patterns_12}.

\begin{table}[htbp]
\centering
\caption{Template injection and L1 hallucination co-occurrence (n=240).}
\label{tab:template_patterns_12}
\begin{threeparttable}
\begin{tabular}{lcc}
\toprule
Model & With Template Patterns & Without Template Patterns \\
\midrule
FT-A     & 199/236 (84.3\% L1) &   0/4 (0.0\% L1)  \\
FT-A+B+C & 202/207 (97.6\% L1) & 16/33 (48.5\% L1) \\
\bottomrule
\end{tabular}
\begin{tablenotes}
\footnotesize
\item Template patterns are defined as in Table~\ref{tab:template_injection}: 
``operating cash flow of USD'', ``USD 3.3 billion'', ``USD 1.3 billion'', 
``USD 1011 million'', and related fabricated expressions.
\end{tablenotes}
\end{threeparttable}
\end{table}

The correlation between template presence and hallucination is striking:

\begin{enumerate}
    \item \textbf{FT-A: near-perfect correlation. }Among the 236 outputs containing template patterns, 84.3\% exhibit L1 hallucination. The 4 outputs without detected templates show zero L1 hallucination. This near-perfect asymmetry (84.3\% vs 0.0\%) is consistent with template injection as the dominant mechanism underlying FT-A hallucination, though we note that the without-template group is small (n=4) and this inference should be treated with appropriate caution.
    \item \textbf{FT-A+B+C: saturated hallucination beyond template patterns. }Only 33 outputs lack detected template patterns, yet even these exhibit 48.5\% L1 hallucination—substantially higher than the corresponding FT-A rate (0.0\%). This suggests that FT-A+B+C has internalized fabrication behavior to a degree that extends beyond the identifiable template patterns captured in Table~\ref{tab:template_injection}: novel fabricated values not in our pattern list may be generated without explicit template triggers. The 207 outputs containing templates show 97.6\% L1 hallucination, approaching complete saturation.
\end{enumerate}

\subsubsection{Qualitative Observation: Repetition Degeneration in 
FT-A$\times$3.5.}

Qualitative inspection of FT-A$\times$3.5 outputs revealed a 
distinct failure mode absent from other model variants: repetitive 
degeneration, in which the model reproduces the same sentence 
pattern continuously until the output length limit is reached. 
This phenomenon occurred in 100\% of synthetic inputs 
(27/27 samples across S0\_Synth and S1\_Synth subsets) but was 
largely absent from real-world inputs (occasional in S0\_Real, 
negligible in S1\_Real).

Crucially, the repeated content is predominantly 
\emph{grounded} in the input --- the model faithfully reproduces 
source numerical values but fails to terminate generation 
appropriately. This distinguishes repetition degeneration from 
template injection: where FT-A and FT-A+B+C fabricate plausible 
but ungrounded values, FT-A$\times$3.5 reproduces grounded values 
but loses generative control.

We attribute this pattern to over-fitting of the output format 
induced by the larger Category A training volume (3,500 samples 
vs.\ 1,000 for FT-A): the model has strongly internalized the 
sentence structure of synthetic financial summaries but has not 
acquired reliable stopping criteria for naturalistic inputs. 
The concentration of repetition in synthetic subsets --- where 
eval inputs closely resemble training data --- is consistent 
with this interpretation. This failure mode can be mitigated 
at inference time via repetition penalty parameters 
\citep{keskarCTRL2019} without retraining, and is orthogonal 
to the hallucination behaviour analysed in the main evaluation.

\subsubsection{Summary of Sensitivity Findings}
Our sensitivity analyses yield three robust conclusions:

\begin{enumerate}
    \item \textbf{Hallucination is not an artifact of input characteristics. }Fine-tuned models exhibit high hallucination rates across data sources and input lengths. FT-A and FT-A+B+C maintain elevated L1 hallucination rates (58.8–100\%) regardless of whether inputs are short structured tables or longer narrative text, and regardless of whether they are drawn from synthetic or real-world sources. The restraint gap persists under all analyzed conditions.
    \item \textbf{Real-world inputs amplify hallucination in models with partial restraint. }Authentic 10-K excerpts trigger substantially higher L1 hallucination than synthetic inputs in FT-A (58.8\% synthetic vs 90.0\% real-world under S1 conditions), suggesting that authentic financial discourse activates stronger canonical narrative priors. This finding implies that evaluation on synthetic data alone may substantially underestimate real-world deployment risk for models that exhibit partial restraint. FT-A+B+C, which shows near-saturated rates on both sources (95.0\% vs 85.0\%), leaves little room for further amplification.
    \item \textbf{Template injection is strongly associated with hallucination. }The near-perfect asymmetry between template presence and L1 hallucination in FT-A (84.3\% with templates vs 0.0\% without, n=236 and n=4 respectively) is consistent with template injection as the dominant pattern underlying FT-A hallucination. For FT-A+B+C, even outputs lacking detected template patterns exhibit 48.5\% L1 hallucination (n=33), suggesting that fabrication behavior has been internalized beyond identifiable surface patterns. Together, these results support template injection as a primary correlate of numerical hallucination in fine-tuned models.
\end{enumerate}

These findings strengthen confidence in our core conclusions while highlighting the importance of evaluating financial language models on authentic domain data. In particular, the amplification of hallucination on real-world inputs underscores the risk of over-relying on synthetic evaluation benchmarks in assessing the reliability of domain-adapted summarization systems.

\subsection{Summary of Key Findings}

Our evaluation reveals five critical findings with direct implications for the deployment and evaluation of financial language models:

\begin{enumerate}
    \item \textbf{Fine-tuning substantially degrades numerical restraint at all detectability levels. }Both FT-A (82.5\%) and FT-A+B+C (90.8\%) exhibit catastrophic L1 hallucination rates, contradicting any assumption that currency-formatted fabrication remains rare in domain-adapted models.
    \item \textbf{Numeracy supervision amplifies rather than mitigates hallucination. }FT-A+B+C exhibits higher hallucination rates than FT-A across all three detectability levels (+7.9pp L1, +11.2pp L2, +9.5pp L3). Improved arithmetic competence does not translate to improved numerical discipline.
    \item \textbf{Weak numerical grounding (S0) represents a catastrophic vulnerability. }Under S0 conditions, fine-tuned models exhibit near-universal L1 hallucination (90–100\%), while the Base model maintains near-complete restraint (0.8\%).
    \item \textbf{Template injection is a primary correlate of hallucination. }Fine-tuned models insert memorized canonical values (e.g., "USD 3.3 billion", "USD 1011 million", "growth of 11\%") regardless of input content, with template presence strongly associated with L1 hallucination (84.3\% in FT-A, 97.6\% in FT-A+B+C).
    \item \textbf{The Base model demonstrates that restraint is achievable. }Near-zero hallucination rates in the Base model confirm that numerical discipline is not inherently beyond model capability—it is specifically degraded through domain fine-tuning.
\end{enumerate}
These findings fundamentally reframe the challenge of numerical hallucination in financial summarization: the problem is not insufficient numerical reasoning, but the loss of numerical restraint through domain adaptation.

\section{Discussion}
\label{sec:discussion}
This study provides a comprehensive analysis of numerical hallucination in financial summarization under domain-specific fine-tuning. Our findings fundamentally challenge prevailing assumptions about numerical hallucination in domain-adapted language models and reveal critical implications for both evaluation methodology and deployment practice.
\subsection{The degradation of Numerical Restraint}
The central finding of this study is that domain fine-tuning substantially degrades numerical restraint---the ability to refrain from numerical generation when grounding is insufficient. This degradation is consistent across all detectability levels.

The Base model demonstrates that numerical restraint is not beyond model capability---it achieves near-zero hallucination across all conditions. This restraint is specifically destroyed through domain adaptation, not inherently absent from the underlying architecture.
This finding inverts the conventional framing of numerical hallucination as a capability deficit. The problem is not that models lack the ability to generate correct numbers; the problem is that fine-tuning removes the barriers preventing models from generating numbers when they should abstain.

\subsection{Template Injection as a Hallucination Mechanism}

Our qualitative analysis reveals a specific mechanism underlying numerical hallucination in fine-tuned models: template injection. Rather than generating numbers through reasoning about input content, fine-tuned models insert memorized canonical values regardless of context.

The evidence for template injection is compelling:

\begin{enumerate}
    \item \textbf{Identical values across unrelated inputs: }The phrase "operating cash flow of USD" appears in 87-98\% of fine-tuned outputs, regardless of whether the source mentions operating cash flow.
    \item \textbf{Model-specific templates: }FT-A favors "USD 3.3 billion" (37\% frequency) while FT-A+B+C favors "USD 1011 million" (50\% frequency), suggesting distinct template libraries learned during their respective training phases.
    \item \textbf{Value distortion:} Source percentages are systematically replaced with template values. The NVIDIA example demonstrates 142\% growth being replaced with 11-14\%—an order of magnitude distortion.
\end{enumerate}

Template injection explains why L1 (Overt) hallucination is so prevalent: fine-tuned models have learned that professional financial summaries should contain specific types of numerical content (revenue, cash flow, growth rates), and they generate canonical examples of these types regardless of input evidence.

\subsection{The Summarization Prior Shift}
Our findings indicate that domain fine-tuning appears to shift the learned objective of summarization:

\textbf{Base Model Objective:} Evidence-preserving compression. The model faithfully reproduces and paraphrases information present in the source, avoiding content that extends beyond provided evidence.

\textbf{Fine-tuned Model Objective:} Canonical narrative generation. The model produces text that matches the expected form and content of professional financial summaries, regardless of whether that content is grounded in the source.

This shift explains the divergent performance observed in Table~\ref{tab:language_fta}: fine-tuned models excel at structured extraction tasks (key points, risk factors) where the expected output format is well-defined, but fail catastrophically at summarization where the expected output is a "complete" financial narrative.

The summarization prior shift has a specific consequence for numerical content: fine-tuned models have learned that financial summaries should contain revenue figures, growth rates, and cash flow statements. When the source lacks these elements, the model generates them anyway---not through reasoning, but through pattern completion based on learned narrative priors.

\subsection{How Numeracy Supervision Relates to Amplified Hallucination}
A counterintuitive finding of this study is that numeracy supervision amplifies hallucination rather than mitigating it. FT-A+B+C exhibits higher hallucination rates than FT-A across all three detectability levels, despite demonstrating improved arithmetic competence under constrained evaluation.

We attribute this effect to the interaction between two factors:

\begin{enumerate}
    \item \textbf{Increased numerical fluency:}Numeracy training improves the model's ability to generate syntactically correct, professionally-formatted numerical expressions. This fluency enables more confident and extensive numerical generation.
    \item \textbf{Unchanged restraint mechanisms: } Numeracy training does not include signals indicating when numerical generation should be avoided. The model learns to compute better but not to abstain when computation is unwarranted.
\end{enumerate}

The result is a model that generates numerical content more fluently and more frequently, without corresponding improvement in grounding discipline. Enhanced competence without enhanced restraint produces more sophisticated fabrication.

This finding has direct practical implications: \textbf{numeracy enhancement should not be applied to summarization systems without accompanying grounding or abstention mechanisms.}

To directly test whether the volume confound accounts for the observed 
hallucination increase, we conducted a volume-matched ablation experiment 
(Section~3.2.4, Table~\ref{tab:ablation_volume}). The ablation model 
FT-A$\times$3.5, trained on 3,500 Category A samples (matching the volume 
of FT-A+B+C but using only financial language data), exhibits \emph{lower} 
L1 hallucination than FT-A (64.6\% vs.\ 82.5\%, $-$17.9pp). In contrast, 
FT-A+B+C—trained on the same volume but with numeracy content added—reaches 
90.8\% ($+$26.2pp above FT-A$\times$3.5). The opposing directions of these 
two effects rule out training volume as a confounding explanation and confirm 
that numeracy supervision, rather than additional data exposure, drives the 
hallucination amplification observed in FT-A+B+C.

\subsection{The Restraint Gap}
Our analysis reveals a fundamental gap between two capabilities that are often conflated:

\begin{itemize}
    \item \textbf{Numerical Competence: }The ability to perform correct numerical computations when explicitly required and when grounding is unambiguous.
    \item \textbf{Numerical Discipline: }The tendency to abstain from absolute numerical generation when input grounding is absent, operationally measured as L1 hallucination rate under S0 conditions (see Section 1.2).
\end{itemize}

\begin{figure}[h]
\centering
\includegraphics[width=0.8\columnwidth]{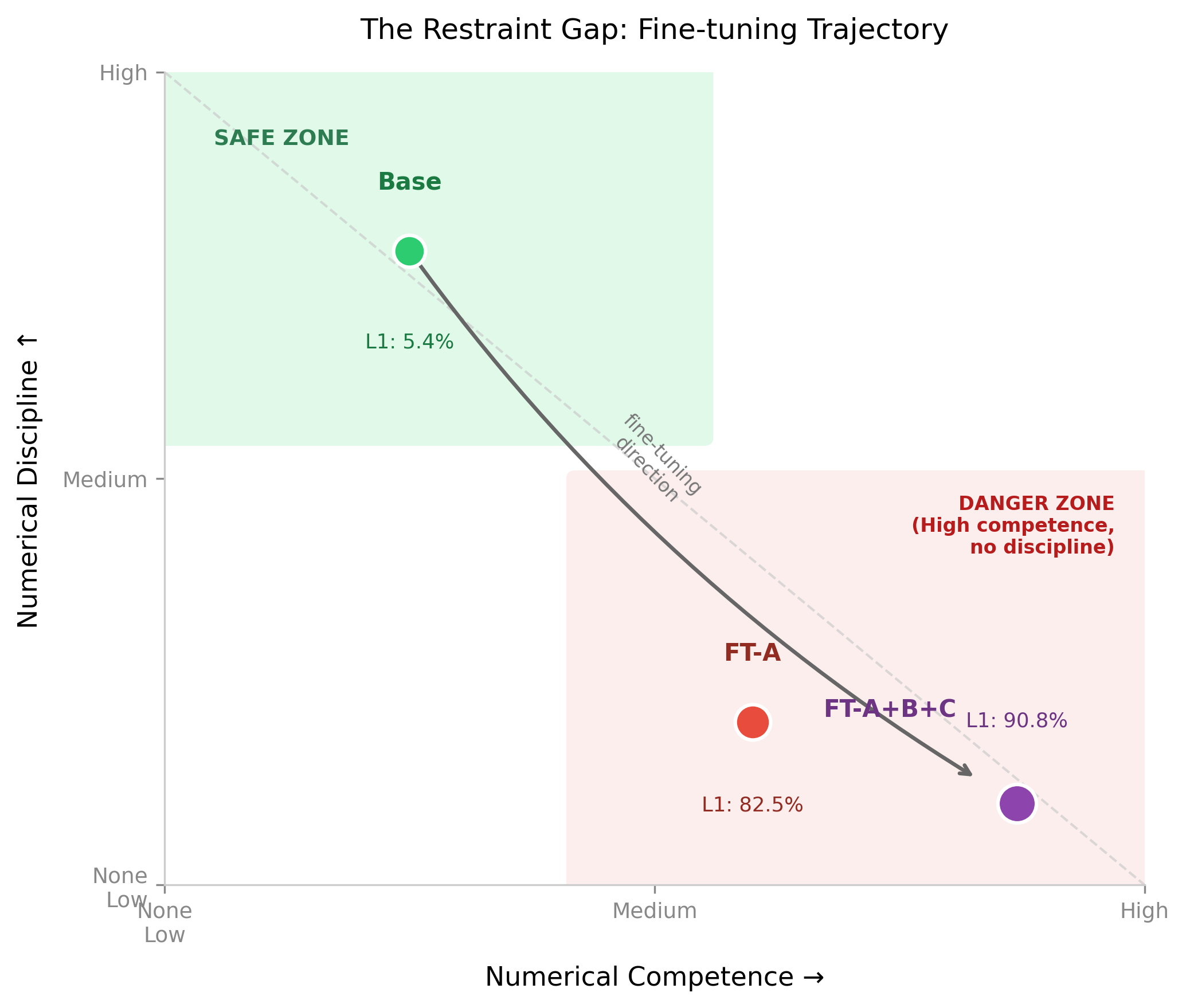}
\caption{The restraint gap: fine-tuning trajectory. Domain adaptation moves models toward higher competence but lower discipline, into the "danger zone" of fluent fabrication. L1 hallucination rates: Base 5.4\%, FT-A 82.5\%, FT-A+B+C 90.8\%.}
\label{fig:restraint}
\end{figure}

Figure~\ref{fig:restraint} illustrates the conceptual 
relationship between competence and discipline.

The restraint gap explains the paradox at the heart of our findings: models can simultaneously improve at arithmetic tasks while becoming dramatically worse at summarization reliability. Competence and discipline are orthogonal dimensions, and current fine-tuning practices optimize the former while destroying the latter.

\subsection{Three distinct failure modes}
Our experiments reveal three distinct failure modes in 
fine-tuned financial summarisation models, each representing 
a different manifestation of degraded generation reliability:

\begin{itemize}
    \item \textbf{Template Injection} (FT-A, FT-A+B+C): 
    Memorised canonical values are inserted regardless of 
    input content, producing fluent but fabricated numerical 
    statements. This is the primary failure mode associated 
    with content-type fine-tuning.
    
    \item \textbf{Repetition-Induced Omission} (FT-A$\times$3.5): 
    Excessive training volume on syntactically uniform data 
    induces the model to repeat a single sentence pattern until 
    the output budget is exhausted, omitting large portions of 
    the input. Hallucination rates appear low but reflect 
    omission rather than restraint.
    
    \item \textbf{Format-Matching Artefact} (Base, S1\_Real): 
    Faithful reproduction of input numerical values using 
    currency-denominated expressions triggers automatic L1 
    detection despite being fully grounded. This is a 
    detection artefact rather than a genuine failure mode.
\end{itemize}

Together, these failure modes illustrate that neither 
training volume nor content type alone produces a reliable 
financial summarisation model. The Base model's maintained 
restraint --- achieved without any domain fine-tuning --- 
remains the strongest evidence that numerical discipline 
is an emergent property of instruction tuning that domain 
adaptation systematically degrades.

\subsection{Implications for Evaluation Methodology}
Our three-level detectability taxonomy reveals systematic blind spots in common evaluation practices:

\begin{itemize}
    \item \textbf{L1-only evaluation }(typical "strict" metrics): Catches currency-formatted fabrication but misses professional-convention numbers and implicit claims. Prior to this study, L1-only evaluation would have suggested FT-A and FT-A+B+C are catastrophically unreliable—a correct conclusion, but one that might have been missed if we had assumed (incorrectly) that strict hallucination rates were near zero.
    \item \textbf{L1+L2 evaluation: }Catches explicit numerical fabrication but misses semantic claims like "maintained stable debt levels." For models with high L3 rates, this underestimates total hallucination.
    \item \textbf{Full L1+L2+L3 evaluation: }Provides comprehensive coverage but requires semantic understanding for L3 detection, making fully automated evaluation challenging.
\end{itemize}

We recommend that evaluation protocols for financial summarization explicitly report hallucination rates at each detectability level, enabling transparent comparison of model reliability across different evaluation stringencies.

\subsection{Practical Deployment Recommendations}
\label{sec:deployment}

Our findings have direct implications for deploying financial language models as components in expert systems. We provide the following recommendations organized by deployment phase.

\subsubsection{Pre-Deployment Evaluation}

\textbf{Recommendation 1: Multi-Level Hallucination Assessment.}
Evaluate candidate models using the L1/L2/L3 detectability taxonomy rather than single-metric assessments. A model appearing reliable under L1-only evaluation may exhibit severe L2/L3 hallucination. Table~\ref{tab:thresholds} provides suggested minimum thresholds for deployment consideration.

\begin{table}[h]
\centering
\caption{Recommended hallucination thresholds for deployment}
\label{tab:thresholds}
\begin{tabular}{@{}lll@{}}
\toprule
\textbf{Evaluation Level} & \textbf{Detection Method} & \textbf{Threshold} \\
\midrule
L1 (Overt) & Currency pattern matching & $<5\%$ \\
L2 (Covert-Explicit) & Number extraction + grounding & $<10\%$ \\
L3 (Covert-Implicit) & Semantic claim verification & $<15\%$ \\
\bottomrule
\end{tabular}
\end{table}

\textbf{Recommendation 2: Grounding Condition Stress Testing.}
Explicitly test model behavior under S0 (weak grounding) conditions. If hallucination rates exceed 20\% under S0, the model is unsuitable for inputs lacking absolute numerical anchors.

\textbf{Recommendation 3: Template Injection Audit.}
Before deployment, analyze model outputs for recurring fabricated values. If specific numerical patterns appear in $>10\%$ of outputs regardless of input content, template injection is present and requires mitigation.

\subsubsection{Runtime Safeguards}

\textbf{Recommendation 4: Grounding Condition Detection.}
Implement automatic classification of inputs into S0/S1 categories. For S0 inputs, route to Base model or apply additional safeguards. Algorithm~\ref{alg:grounding} provides a reference implementation.

\begin{algorithm}[h]
\caption{Grounding Condition Classification}
\label{alg:grounding}
\begin{algorithmic}[1]
\Require Input text $T$
\Ensure Grounding condition $\in \{S0, S1\}$
\State Define pattern $P \leftarrow$ \texttt{r'\textbackslash\$[\textbackslash d,]+|\textbackslash d+\textbackslash s*(million|billion|USD)'}
\If{$\text{match}(P, T) \neq \emptyset$}
    \State \Return $S1$ \Comment{Strong grounding}
\Else
    \State \Return $S0$ \Comment{Weak grounding}
\EndIf
\end{algorithmic}
\end{algorithm}

\textbf{Recommendation 5: Template Detection Filter.}
Implement post-generation filtering for known template patterns. Outputs containing detected templates should be flagged for human review. Based on our analysis, the following patterns indicate template injection with high probability:

\begin{itemize}
    \item \texttt{USD 3.3 billion} (43\% frequency in FT-A)
    \item \texttt{USD 1011 million} (58\% frequency in FT-A+B+C)
    \item \texttt{operating cash flow of USD} (87--98\% frequency)
    \item \texttt{growth of 11\%} (10--25\% frequency)
\end{itemize}

\textbf{Recommendation 6: Confidence-Based Suppression.}
When available, use model confidence scores to suppress low-confidence numerical content. Numerical tokens generated with confidence below a threshold (e.g., 0.8) should trigger review flags.

\subsubsection{Architectural Recommendations}

\textbf{Recommendation 7: Hybrid Architecture.}
Consider hybrid architectures that leverage different models for different capabilities, as illustrated in Table~\ref{tab:hybrid}.

\begin{table}[h]
\centering
\caption{Hybrid architecture for financial summarization}
\label{tab:hybrid}
\begin{tabular}{@{}lll@{}}
\toprule
\textbf{Component} & \textbf{Model} & \textbf{Rationale} \\
\midrule
Numerical content & Base model & Maintains restraint \\
Stylistic enhancement & Fine-tuned model & Domain fluency \\
Final output & Merger/selector & Best of both \\
\bottomrule
\end{tabular}
\end{table}

\textbf{Recommendation 8: Retrieval Augmentation.}
Implement retrieval-augmented generation (RAG) to anchor numerical content to verifiable source passages. This may suppress template injection by providing explicit grounding signals at generation time.

\subsubsection{Monitoring and Maintenance}

\textbf{Recommendation 9: Continuous Hallucination Monitoring.}
Deploy ongoing monitoring for hallucination indicators:
\begin{itemize}
    \item Track template pattern frequency over time
    \item Monitor L1/L2 rates on production samples
    \item Alert on sudden increases in numerical generation density
\end{itemize}

\textbf{Recommendation 10: User Feedback Integration.}
Implement feedback mechanisms allowing users to flag suspected hallucinations. Aggregate feedback to identify emerging template patterns not captured in initial audits.


\subsection{Limitations of This Analysis}

Several limitations should be considered when interpreting our findings:

\begin{enumerate}
    \item \textbf{Detection methodology: }Our automatic detection of L2 and L3 hallucination relies on pattern matching and keyword detection, which may miss subtle cases or produce false positives. Manual verification of a subset confirms the overall patterns but cannot guarantee perfect accuracy.
    \item \textbf{Model scale: }Our experiments use 7B-parameter models. Larger models may exhibit different hallucination dynamics, potentially with greater restraint or more sophisticated fabrication. It should be noted, however, that 7B-parameter models represent the most widely deployed scale for domain-specific fine-tuning in practice, and our findings have direct implications for this deployment tier regardless of how larger models may behave.
    \item \textbf{Training data composition: }The specific hallucination patterns observed (e.g., template values like "USD 3.3 billion") depend on training data composition. Different training corpora would produce different template libraries.
    \item \textbf{Task scope: }We focus on financial summarization. Other financial NLP tasks (sentiment analysis, question answering, extraction) may exhibit different hallucination characteristics.
\end{enumerate}

\subsection{Summary}
Our analysis demonstrates that numerical hallucination in financial summarization is primarily associated with the substantial degradation of numerical restraint through domain fine-tuning, rather than by insufficient numerical reasoning capability. Fine-tuned models exhibit catastrophic hallucination rates at all detectability levels, with numeracy supervision amplifying rather than mitigating the problem.

The path toward reliable financial summarization requires a fundamental shift from optimizing numerical competence to enforcing numerical discipline. This will require explicit training for restraint, grounding-aware generation mechanisms, and evaluation protocols that assess hallucination across all detectability levels.

\section{Limitations and Future Work}
\label{sec:limitations}

This study has several limitations that point to important directions for future research and practical system design.

\subsection{Detection Methodology Limitations}

Our three-level detectability taxonomy (L1/L2/L3) relies on pattern matching and heuristic criteria for automatic hallucination detection. While L1 (Overt) detection through currency-symbol matching is highly reliable, L2 (Covert-Explicit) and L3 (Covert-Implicit) detection face inherent challenges.

For L2, our approach extracts numerical expressions and cross-references against source text, but may miss cases where fabricated values coincidentally match unrelated numbers in the input. For L3, semantic pattern matching for implicit claims (e.g., "maintained stable debt") cannot capture all forms of ungrounded quantitative implication.

Developing more principled detection methods---potentially incorporating structured reasoning traces, entailment models, or hybrid human-machine evaluation---remains an open challenge. Such advances would enable more precise quantification of hallucination rates, particularly at the L3 level where semantic understanding is required.

Recent work on adaptive evaluation frameworks offers a 
complementary perspective on this limitation. Ding et al.\ 
\citep{ding2026adarubric} propose ADARUBRIC, a task-adaptive 
rubric generation approach with confidence-weighted scoring 
that dynamically adjusts evaluation dimensions to the specific 
task instance. Applied to numerical hallucination detection, 
such an approach could in principle generate instance-specific 
grounding criteria --- for example, weighting numerical 
precision more heavily for inputs containing explicit 
calculations, and semantic consistency more heavily for 
narrative-only inputs. We view adaptive rubric generation as 
a promising direction for extending the L1/L2/L3 taxonomy 
toward more flexible, task-conditioned evaluation, and leave 
this integration to future work.

\subsection{Template Injection Mechanism}

Our analysis identifies template injection as a primary correlate of hallucination, evidenced by recurring fabricated values across unrelated inputs (e.g., "USD 3.3 billion" in 37\% of FT-A outputs). However, our understanding of this mechanism remains descriptive rather than explanatory.

Future work should investigate:

\begin{itemize}
    \item \textbf{Where templates are learned: }Whether specific training examples or corpus statistics drive template formation
    \item \textbf{How templates are triggered: }What input features activate template insertion versus faithful generation
    \item \textbf{Whether templates can be suppressed: }Through targeted fine-tuning, activation editing, or constrained decoding
\end{itemize}

Understanding the mechanistic basis of template injection could enable targeted interventions that preserve domain fluency while eliminating fabrication.

\subsection{Training Objectives for Numerical Restraint}

Our analysis focuses on supervised fine-tuning with static instruction-following objectives. While this reflects common engineering practice, supervised fine-tuning does not explicitly model the decision of whether numerical generation should occur---it only models what to generate given that generation will occur.

The restraint gap we identify---between numerical competence and numerical discipline---suggests that fundamentally different training signals are required. Future work could explore:

\begin{itemize}
    \item \textbf{Reinforcement learning with restraint rewards: }Explicitly penalizing ungrounded numerical generation while rewarding appropriate abstention. However, recent work has shown that reward model overoptimization can introduce new failure modes, suggesting that restraint-aware objectives must be carefully designed.
    \item \textbf{Contrastive training with abstention examples: }Including negative examples that demonstrate appropriate numerical abstention under weak grounding (S0) conditions, paired with positive examples of faithful reproduction under strong grounding (S1) conditions.
    \item \textbf{Confidence-calibrated generation: }Training models to output uncertainty estimates alongside numerical content, enabling downstream suppression of low-confidence fabrication.
\end{itemize}

As demonstrated in our results, numeracy supervision alone improves arithmetic competence but substantially degrades numerical restraint. Future training approaches must explicitly optimize for discipline alongside competence.

\subsection{Mitigation Strategies}

Our experiments are conducted using relatively small-scale, cost-effective fine-tuning configurations. While this choice reflects realistic deployment constraints and strengthens the applicability of our findings, it limits exploration of mitigation strategies. Promising directions for restraint-preserving deployment include:

\begin{itemize}
    \item \textbf{Retrieval-Augmented Generation. }Conditioning generation on retrieved source passages may anchor numerical content to verifiable evidence, potentially suppressing template injection. Recent work has demonstrated that RAG-based approaches substantially reduce hallucination in knowledge-intensive tasks by grounding generation in retrieved context rather than parametric memory\cite{lewis2020retrieval}. In the financial summarization setting identified in this study, RAG could be particularly effective under S0 conditions, where the absence of numerical anchors in the input creates the highest vulnerability to fabrication. By retrieving relevant numerical facts from an external knowledge base at inference time, a RAG system could supply the grounding evidence that the input alone lacks, potentially converting an effective S0 condition into an S1 condition without retraining.

    \item \textbf{Instruction Contrastive Decoding.}
Wang et al.\ \citep{wang-etal-2024-mitigating} propose Instruction Contrastive 
Decoding (ICD), which mitigates hallucination by contrasting 
output distributions under standard and ``disturbance'' 
instructions. The disturbance instruction is designed to 
amplify the model's language priors, and the difference between 
standard and disturbance distributions is used to suppress 
prior-driven generation while preserving evidence-based content.

The template injection mechanism identified in this paper 
represents precisely the kind of language-prior-driven 
fabrication that ICD targets. In our setting, a disturbance 
prompt could be constructed by instructing the fine-tuned model 
to generate a financial summary \emph{emphasising} specific 
numerical values --- this would amplify the template injection 
behaviour (e.g., increasing the probability of ``USD 3.3 
billion''-type outputs), whose distribution could then be 
subtracted from the standard output distribution at decoding 
time. The resulting distribution would suppress memorised 
templates while preserving input-grounded numerical content.

Formally, let $p_\theta(y \mid x, i_\text{std})$ denote the 
output distribution under the standard summarisation instruction 
and $p_\theta(y \mid x, i_\text{dist})$ under a disturbance 
instruction designed to elicit template injection. ICD defines 
the adjusted logit as:
\begin{equation}
\text{logit}_\text{ICD}(y_t) = \text{logit}(y_t \mid x, 
i_\text{std}) - \alpha \cdot \text{logit}(y_t \mid x, 
i_\text{dist})
\end{equation}
where $\alpha$ controls the degree of contrast. Applied to 
S0 inputs --- where template injection is near-universal 
(96.7\% L1 for FT-A) --- ICD could directly target the 
restraint gap without requiring model retraining. Validating 
this approach empirically is an important direction for the 
planned follow-on study on hallucination mitigation.
    
    \item \textbf{Grounding-aware decoding. }Constrained decoding strategies that suppress numerical token generation when source grounding is insufficient may directly address the restraint gap identified in this study. Recent approaches to factuality-aware decoding\cite{shi2024trusting,chuang2024dola} demonstrate that inference-time constraints can reduce hallucination without modifying model weights. Applied to the template injection mechanism we identify, a decoding constraint that monitors the source input for numerical anchors and suppresses currency-denominated token sequences when none are present could directly target L1 hallucination under S0 conditions. Such an approach would preserve the domain fluency of fine-tuned models while restoring numerical discipline at the generation level.
    \item \textbf{Post-generation verification.}Automated fact-checking pipelines that detect and flag template injection patterns before output delivery represent a practical near-term mitigation. Our identification of specific high-frequency template patterns (e.g., "operating cash flow of USD" appearing in 80–97\% of fine-tuned model outputs, "USD 3.3 billion" in 37\% of FT-A outputs, and "USD 1011 million" in 50\% of FT-A+B+C outputs) suggests that a lightweight rule-based filter targeting these patterns could eliminate a substantial proportion of L1 hallucinations with minimal computational overhead. More sophisticated approaches using natural language inference models to verify numerical claims against source passages\cite{min2023factscore} could further address L2 and L3 hallucination. Importantly, post-generation verification does not require model retraining and can be deployed as a wrapper around existing fine-tuned systems.
    \item \textbf{Hybrid architectures. }Routing numerical content generation to the Base model (which maintains restraint) while using fine-tuned models for stylistic enhancement remains a practical deployment option. Our finding that the Base model maintains near-zero L1 hallucination (5.4\% overall, 0\% under S0 synthetic conditions) while fine-tuned models exhibit catastrophic rates (82.5–90.8\%) suggests a clear functional division: fine-tuned models can be used for structural formatting, terminology alignment, and professional tone, while numerical content is either passed through directly from the source or generated by the restrained Base model. This hybrid approach exploits the complementary strengths identified in our restraint gap analysis without requiring new training.
    \item \textbf{A key insight from our findings }is that these mitigation strategies are not mutually exclusive and may be most effective in combination. Grounding condition detection (Algorithm 1) can serve as an upstream router: S0 inputs are redirected to the Base model or a RAG-augmented pipeline, while S1 inputs are processed by the fine-tuned model with post-generation verification. This layered approach directly addresses the vulnerability profile revealed by our S0/S1 stratification.
\end{itemize}

Investigating whether these approaches can suppress template injection without sacrificing domain fluency represents a critical direction for practical deployment.

\subsection{Model Scale and Architecture}
Our experiments use Mistral-7B-Instruct-v0.2 as the base model, 
reflecting the most widely deployed scale for domain-specific 
financial fine-tuning in resource-constrained settings. While 
this choice enables controlled, reproducible experimentation, 
the generalisability of our findings to larger models warrants 
explicit discussion.

\paragraph{Potential for greater restraint at larger scale}
Larger models (e.g., 13B, 70B, or frontier-scale) may exhibit 
stronger baseline numerical discipline due to more robust 
instruction-following capabilities and broader exposure to 
diverse numerical contexts during pretraining. If larger models 
maintain stronger priors toward evidence-based generation, 
domain fine-tuning may induce a smaller restraint gap than we 
observe at 7B scale. This possibility suggests that the 
hallucination rates we report (82.5--90.8\% L1 for fine-tuned 
models) may represent an upper bound on severity, with larger 
models potentially exhibiting more moderate degradation.

\paragraph{Potential for more sophisticated fabrication}
Conversely, larger models may generate more contextually 
plausible fabrications that are harder to detect under the 
L1/L2/L3 taxonomy. If template injection at larger scale 
produces values that are numerically plausible given the input 
domain (e.g., revenue figures consistent with the company's 
industry and size), L2 and L3 detection rates may 
underestimate true hallucination risk. In this scenario, the 
restraint gap would be more dangerous rather than less severe, 
as fabrications would be harder to identify through automated 
means. Recent findings on frontier models suggest that 
instruction-following improvements do not uniformly translate 
to factual grounding in domain-specific generation tasks 
\citep{huang2023survey}, lending plausibility to this concern.

\paragraph{Scale-dependent failure modes}
Our volume-matched ablation (Section~3.2.4) reveals that 
training data volume introduces its own failure modes 
independent of model scale: FT-A$\times$3.5 exhibits 
repetitive degeneration and omission-by-repetition rather 
than template injection. This suggests that the relationship 
between training scale, model scale, and hallucination 
behaviour is non-monotonic and warrants systematic 
cross-scale investigation.

\paragraph{Practical relevance of 7B-scale findings}
Despite these caveats, our findings at 7B scale have direct 
practical relevance. Models in the 7B parameter range are 
among the most commonly deployed for domain-specific financial 
applications due to their inference efficiency and fine-tuning 
accessibility. The restraint gap framework and L1/L2/L3 
evaluation taxonomy we introduce are scale-agnostic: they can 
be applied to evaluate numerical discipline at any model scale, 
providing a methodological foundation for future cross-scale 
comparisons. We leave systematic cross-scale analysis to future 
work, noting that our taxonomy and experimental protocol are 
designed to facilitate such comparisons.

\subsection{Domain Generalization}

Our study focuses on financial summarization, but the observed failure modes—template injection, restraint degradation, and the competence-discipline gap—are unlikely to be domain-specific. Canonical narrative priors may arise in other high-stakes domains:

\begin{itemize}
    \item \textbf{Healthcare: }Clinical summaries may exhibit template injection of standard vital signs, lab values, or treatment protocols
    \item \textbf{Legal:}Case summaries may fabricate canonical precedent citations or statutory references
    \item \textbf{Scientific reporting:}Research summaries may inject template statistics or p-values
\end{itemize}

Extending the L1/L2/L3 detectability framework to these domains would test the generality of our findings and inform domain-specific evaluation protocols. The restraint gap may prove to be a universal consequence of domain fine-tuning in professional contexts where structured quantitative content is expected.

\subsection{Real-World Deployment Validation}

Our evaluation uses a curated dataset of 240 samples spanning synthetic and real-world financial texts. While this enables controlled analysis, production deployment involves:

\begin{itemize}
    \item Higher volume and greater input diversity
    \item Adversarial or unusual inputs that may trigger unexpected template activation
    \item User reliance on outputs for consequential decisions
\end{itemize}

Future work should validate our findings through larger-scale deployment studies, potentially incorporating user feedback on output reliability and downstream decision quality.

\subsection{Summary}

Overall, this work suggests that improving numerical reliability in domain-adapted language models requires a fundamental shift from optimizing numerical competence toward enforcing numerical discipline. The restraint gap we identify is not merely an evaluation artifact—it reflects a genuine capability distinction that current fine-tuning practices fail to address.

\begin{itemize}
    \item \textbf{Training objectives }that explicitly reward restraint alongside competence
    \item \textbf{Detection methods }that reliably identify hallucination across all detectability levels
    \item \textbf{Mitigation strategies }that suppress template injection without sacrificing domain fluency
    \item \textbf{Evaluation protocols }that assess discipline as a first-class capability
\end{itemize}

We leave this agenda for future investigation, with the hope that our detectability taxonomy and restraint gap framing provide useful foundations for subsequent work.

\section{Conclusion}
\label{sec:conclusion}

This paper presents a systematic analysis of numerical hallucination in financial summarization under domain-specific fine-tuning. Through controlled comparisons across Base, FT-A, and FT-A+B+C models, we demonstrate that numerical hallucination cannot be attributed to insufficient numerical reasoning capability---instead, it arises from the degradation of numerical restraint through domain adaptation.
Our central findings fundamentally challenge prevailing assumptions:

\textbf{Finding 1: Fine-tuning substantially degrades numerical restraint at all detectability levels. } The Base model maintains near-zero hallucination (5.4\% L1, 3.8\% L2, 17.9\% L3\textsuperscript{*}), while FT-A exhibits 82.5\% L1 hallucination and FT-A+B+C reaches 90.8\%. Domain adaptation does not merely shift hallucination from detectable to undetectable forms—it is associated with pervasive fabrication across all evaluation criteria.

\textbf{Finding 2: Numeracy supervision amplifies hallucination. }
Despite improving arithmetic competence under constrained evaluation, FT-A+B+C exhibits higher hallucination rates than FT-A at every detectability level. Enhanced numerical fluency enables more sophisticated fabrication, while the absence of restraint training removes barriers to ungrounded generation.

\textbf{Finding 3: Template injection is a primary hallucination mechanism.} Fine-tuned models insert memorized canonical values (e.g., "USD 3.3 billion", "growth of 11\%") regardless of input content, suggesting that domain adaptation appears to strengthen narrative priors that compete with evidence-based generation.

\textbf{Finding 4: The restraint gap explains the paradox.} Numerical competence (the ability to compute correctly) and numerical discipline (the ability to abstain when grounding is insufficient) are orthogonal capabilities. Current fine-tuning practices optimize competence while destroying discipline, producing models that fabricate more fluently.
We introduce a three-level detectability taxonomy (L1 Overt, L2 Covert-Explicit, L3 Covert-Implicit) that enables systematic analysis of evaluation blind spots. This taxonomy reveals that any evaluation protocol targeting only currency-formatted fabrication will miss substantial hallucination in domain-adapted models.
Our findings have direct implications for deployment: fine-tuned financial models should not be used for summarization without explicit grounding or abstention mechanisms, particularly under weak numerical grounding (S0) conditions where hallucination rates approach 100\%. The Base model's maintained restraint demonstrates that numerical discipline is achievable---future work must focus on preserving or restoring this capability through fine-tuning.
In summary, improving numerical reliability in financial summarization requires a fundamental shift from optimizing numerical accuracy toward enforcing numerical discipline. Bridging this restraint gap will require joint advances in training objectives, decoding strategies, and evaluation protocols that explicitly reward appropriate abstention alongside correct generation.

\section*{Data and Code Availability}
\label{sec: section_reproducable}
All code used for data generation, fine-tuning, inference, and evaluation in this study is publicly available at:

\url{https://github.com/ironwire/mistral_tuning}

Due to licensing constraints, raw financial documents are not redistributed; however, all synthetic data generation scripts, evaluation protocols, and experimental configurations are provided to ensure full reproducibility of the reported results.


\bibliographystyle{elsarticle-num}
\bibliography{EAAI_References}

\appendix 
\section{Training Detail}
\label{sec:Appendix A}
This appendix provides detailed information on model configurations, fine-tuning procedures, and training settings used throughout our experiments.

\subsection{Base Model and Fine-Tuning Method}
All experiments are conducted using Mistral-7B-Instruct-v0.2~\cite{jiang2023mistral} as the base model. Fine-tuning is performed using QLoRA~\cite{dettmers2023qlora} , a parameter-efficient adaptation method that enables low-resource training while preserving the base model weights.
\paragraph{Choice of Parameter-Efficient Method}
We selected QLoRA over alternative parameter-efficient 
fine-tuning methods such as prefix-tuning~\cite{li2021prefix} or adapter layers 
due to its superior performance on generation tasks under resource constraints 
and its ability to preserve full model capacity during inference.
We adopt supervised fine-tuning (SFT) exclusively and do not apply reinforcement learning or preference optimization. This choice reflects common industrial fine-tuning practices and allows for controlled attribution of observed behavioral changes to data composition rather than optimization complexity.

\subsection{Fine-Tuning Configurations}
Three fine-tuned variants are trained:
\begin{itemize}
    \item \textbf{FT-C:} Fine-tuned exclusively on numeracy-focused financial calculation tasks under strong output constraints.
    \item \textbf{FT-A:} Fine-tuned on financial language tasks, including summarization, key point extraction, and risk factor extraction.
    \item \textbf{FT-A+B+C:} Fine-tuned on a mixture of financial language (A), financial knowledge QA (B), and numeracy-focused calculation tasks (C).
\end{itemize}
All models are trained for a single epoch to minimize overfitting and to emphasize behavioral shifts induced by data composition rather than prolonged optimization.

\subsection{Training Hyperparameters}
Unless otherwise specified, all fine-tuning runs use the following settings:
\begin{itemize}
    \item Optimizer: AdamW
    \item Learning rate: 2e-4
    \item Batch size: 1
    \item Gradient accumulation steps: 16
    \item Effective batch size: 16
    \item Maximum sequence length: 1024 tokens
    \item Precision: 4-bit quantization (NF4) with FP16 computation
    \item LoRA rank: 8
    \item LoRA target modules: attention projection layers
\end{itemize}
We adopt QLoRA, which combines 4-bit quantization 
with Low-Rank Adaptation (LoRA)\cite{hu2022lora} to enable efficient fine-tuning.
Training is performed on a single NVIDIA RTX A3000 (12GB) GPU. All runs complete within several hours.
\subsection{Inference Settings}
Unless otherwise stated, inference is performed using greedy decoding with the following parameters:
\begin{itemize}
    \item Temperature: 0.0
    \item Top-p: disabled
    \item Max new tokens: task-dependent (typically 128-- 256)
\end{itemize}
This configuration ensures deterministic outputs and avoids stochastic variation during evaluation.

\section{Dataset Construction}
\subsection{Overview of Data Categories}
We construct three primary categories of training data, all formatted as instruction–input–output triples:
\begin{itemize}
    \item \textbf{Category A (Financial Language Tasks): }
    
Synthetically generated tasks including summarization, key point extraction, and risk factor extraction. Data are designed to improve domain-specific language use and structural conformity while controlling narrative structure and eliminating confounding numerical cues.

    \item \textbf{Category B (Financial Knowledge QA): }

Short-form question–answer pairs covering financial concepts, accounting definitions, statement relationships, and reporting terminology. Answers prioritize conceptual clarity over numerical calculation.

    \item \textbf{Category C (Financial Numeracy Tasks): }

Calculation-focused tasks derived from templated financial tables with known ground-truth values. Tasks include ratio computation, growth rates, and margin analysis, with outputs constrained to numeric formats to improve calculation reliability.

\end{itemize}

\subsection{Real-World Financial Text Evaluation Set}

To complement synthetic data, we construct an evaluation-only dataset consisting of \textbf{80 excerpts extracted from real-world 10-K reports}, evenly split between S0 and S1 conditions. These excerpts are not used for training and serve exclusively to assess generalization to authentic financial disclosures.

The inclusion of real 10-K text enables evaluation of hallucination behavior in realistic reporting contexts while maintaining controlled numerical grounding conditions.

\subsection{Dataset Sizes}
Approximate dataset sizes are as follows:

\begin{itemize}
    \item Category A (training): ~1,000 samples
    \item Category B (training): ~500 samples
\item Category C (training): ~2,000 samples

\item Summarization evaluation set: 240 samples
\begin{itemize}
    \item Synthetic: 160 samples (80 S0 / 80 S1)
    \item Real-world 10-K: 80 samples (40 S0 / 40 S1)
\end{itemize}
\end{itemize}

\section{Supplementary Quantitative Results}
\label{appendix:results}

This appendix provides detailed quantitative results supporting the main findings presented in Section~4. All analyses use the three-level detectability taxonomy (L1/L2/L3) introduced in Section~3.4.

\subsection{Full Hallucination Rate Breakdown}
\label{appendix:breakdown}

Table~\ref{tab:c14_full_breakdown} presents the complete breakdown of hallucination rates by sample category (grounding condition $\times$ data source) and model variant.

\begin{table}[htbp]
\caption{Hallucination counts and rates (\%) by category and model (n=240).}
\label{tab:c14_full_breakdown}
\begin{threeparttable}
\begin{tabular}{llrcccc}
\toprule
Category & Model & $n$ & L1 (Overt) & L2 (Cov-Exp) & L3 (Cov-Imp) & Template \\
\midrule
\multirow{3}{*}{S0\_Synth}
 & Base     & 80 &  0 (0.0\%)   &  0 (0.0\%)   & 33 (41.2\%)\tnote{*} &  0 (0.0\%)   \\
 & FT-A     & 80 & 80 (100.0\%) &  0 (0.0\%)   &  0 (0.0\%)   & 80 (100.0\%) \\
 & FT-A+B+C & 80 & 80 (100.0\%) &  0 (0.0\%)   &  2 (2.5\%)   & 80 (100.0\%) \\
\midrule
\multirow{3}{*}{S0\_Real}
 & Base     & 40 &  1 (2.5\%)   &  0 (0.0\%)   &  4 (10.0\%)  &  0 (0.0\%)   \\
 & FT-A     & 40 & 36 (90.0\%)  &  0 (0.0\%)   &  4 (10.0\%)  & 36 (90.0\%)  \\
 & FT-A+B+C & 40 & 28 (70.0\%)  &  0 (0.0\%)   &  7 (17.5\%)  & 26 (65.0\%)  \\
\midrule
\multirow{3}{*}{S1\_Synth}
 & Base     & 80 &  0 (0.0\%)   &  0 (0.0\%)   &  0 (0.0\%)   &  0 (0.0\%)   \\
 & FT-A     & 80 & 47 (58.8\%)  & 47 (58.8\%)  &  2 (2.5\%)   & 80 (100.0\%) \\
 & FT-A+B+C & 80 & 76 (95.0\%)  & 76 (95.0\%)  & 13 (16.2\%)  & 80 (100.0\%) \\
\midrule
\multirow{3}{*}{S1\_Real}
 & Base     & 40 & 14 (35.0\%)\tnote{\dag}  &  9 (22.5\%)\tnote{\dag} &  6 (15.0\%)  &  0 (0.0\%)   \\
 & FT-A     & 40 & 36 (90.0\%)  & 36 (90.0\%)  &  3 (7.5\%)   & 40 (100.0\%) \\
 & FT-A+B+C & 40 & 34 (85.0\%)  & 34 (85.0\%)  & 10 (25.0\%)  & 21 (52.5\%)  \\
\midrule
\multirow{3}{*}{Overall}
 & Base     & 240 & 15 (5.4\%)   &  9 (3.8\%)   & 43 (17.9\%)  &  0 (0.0\%)   \\
 & FT-A     & 240 & 198 (82.5\%) & 83 (34.6\%)  &  9 (3.8\%)   & 236 (98.3\%) \\
 & FT-A+B+C & 240 & 218 (90.8\%) & 110 (45.8\%) & 32 (13.3\%)  & 207 (86.2\%) \\
\bottomrule
\end{tabular}
\begin{tablenotes}
\footnotesize
\item[*] Base S0\_Synth L3 (33/80, 41.2\%) reflects faithful paraphrase 
of qualitative state descriptions present in the source input rather than 
genuine ungrounded hallucination. See Appendix~D.5.
\item[\dag] Elevated Base S1\_Real L1 and L2 rates are attributable to 
format-matching artefacts: the Base model faithfully reproduces numerical 
values from authentic 10-K excerpts using currency-denominated expressions 
that trigger detection rules despite being fully grounded in the source.
\end{tablenotes}
\end{threeparttable}
\end{table}

Key observations from Table~\ref{tab:c14_full_breakdown}:

\begin{itemize}
    \item \textbf{S0\_Synth}: Fine-tuned models exhibit 100\% L1 hallucination under weak grounding with synthetic data, while Base maintains 0\%. This result is unchanged from the original findings and represents the clearest demonstration of the restraint gap.
    
    \item \textbf{S0\_Real}: Real-world inputs trigger elevated L3 hallucinations in fine-tuned models (FT-A: 10.0\%, FT-A+B+C: 17.5\%). Unlike the original findings, L2 hallucination is 0\% for both fine-tuned models under S0 conditions, consistent with the revised L2 detection protocol which targets absolute magnitude expressions only, excluding percentage-based quantities.
    
    \item \textbf{S1\_Synth}: Even with explicit numerical grounding, FT-A hallucinates in 58.8\% of samples and FT-A+B+C in 95.0\%, both higher than the original findings (50\% and 97.5\% respectively), reflecting the expanded synthetic sample pool.
    
    \item \textbf{S1\_Real}: Real-world grounded inputs show 90.0\% L1 hallucination for FT-A and 85.0\% for FT-A+B+C. Base model shows an elevated rate (35.0\%) attributable to format-matching artefacts rather than genuine hallucination, as discussed in Table~\ref{tab:hallucination_bylevels}.
    
    \item \textbf{Template injection}: Template patterns appear in 86.2–98.3\% of fine-tuned model outputs but 0\% of Base outputs, confirming that template injection is a consequence of fine-tuning rather than an inherent model behavior.
\end{itemize}

\subsection{Hallucination Rate Confidence Intervals}
\label{appendix:ci}

Table~\ref{tab:ci_overall} presents hallucination rates with 95\% Wilson score confidence intervals, providing statistical bounds on the observed rates.

\begin{table}[htbp]
\centering
\caption{Overall hallucination rates with 95\% Wilson score confidence intervals}
\label{tab:ci_overall}
\begin{tabular}{@{}llccc@{}}
\toprule
\textbf{Model} & \textbf{Level} & \textbf{Rate (\%)} & \textbf{95\% CI} & \textbf{n} \\
\midrule
\multirow{3}{*}{Base} 
    & L1 & 5.4  & [3.8\%, 10.1\%]   & 240 \\
    & L2 & 3.8  & [2.0\%, 7.0\%]   & 240 \\
    & L3 & 17.9  & [13.6\%, 23.3\%]   & 240 \\
\midrule
\multirow{3}{*}{FT-A} 
    & L1 & 82.5 & [77.6\%, 87.2\%] & 240 \\
    & L2 & 34.6 & [28.9\%, 40.8\%] & 240 \\
    & L3 & 3.8 & [2.0\%, 7.0\%]  & 240 \\
\midrule
\multirow{3}{*}{FT-A+B+C} 
    & L1 & 90.8 & [86.5\%, 93.9\%] & 240 \\
    & L2 & 45.8 & [39.6\%, 52.2\%] & 240 \\
    & L3 & 13.3 & [9.6\%, 18.2\%] & 240 \\
\bottomrule
\end{tabular}
\end{table}

The confidence intervals confirm that the differences between models are statistically significant. Notably:

\begin{itemize}
    \item Base L1 upper bound (10.1\%) does not overlap with FT-A L1 lower bound (77.6\%), confirming a significant difference (p < 0.001).
    
    \item Although the confidence intervals for FT-A [77.6\%, 87.2\%] and FT-A+B+C [86.5\%, 93.9\%] overlap marginally, the difference is confirmed statistically significant by Fisher's exact test (p = 0.015, Table C.18), providing evidence that numeracy supervision further increases L1 hallucination.
\end{itemize}

Table~\ref{tab:ci_grounding} presents L1 hallucination rates stratified by grounding condition.

\begin{table}[htbp]
\centering
\caption{L1 hallucination rates by grounding condition with 95\% CI}
\label{tab:ci_grounding}
\begin{tabular}{@{}llccc@{}}
\toprule
\textbf{Model} & \textbf{Condition} & \textbf{Rate (\%)} & \textbf{95\% CI} & \textbf{n} \\
\midrule
\multirow{2}{*}{Base} 
    & S0 & 0.8  & [0.1\%, 4.6\%]    & 120 \\
    & S1 & 11.7  & [7.1\%, 18.6\%]   & 120 \\
\midrule
\multirow{2}{*}{FT-A} 
    & S0 & 96.7 & [91.7\%, 98.7\%] & 120 \\
    & S1 & 69.2  & [60.4\%, 76.7\%]  & 120 \\
\midrule
\multirow{2}{*}{FT-A+B+C} 
    & S0 & 90.0 & [83.3\%, 94.2\%]  & 120 \\
    & S1 & 91.7 & [85.3\%, 95.4\%]  & 120 \\
\bottomrule
\end{tabular}
\end{table}

The grounding condition analysis reveals:

\begin{itemize}
    \item \textbf{FT-A shows grounding sensitivity}: SS0 (96.7\%) vs S1 (69.2\%) indicates a partial protective effect of numerical grounding, with a difference of 27.5 percentage points. This suggests that explicit numerical anchors in S1 inputs provide some constraint on fabrication in FT-A, though hallucination remains catastrophically high in both conditions.
    
    \item \textbf{FT-A+B+C eliminates grounding protection}: S0 (90.0\%) and S1 (91.7\%) show near-identical L1 hallucination rates, confirming that numeracy supervision removes even the partial restraint mechanisms observed in FT-A.
    
    \item \textbf{Base maintains restraint in both conditions}: S0 upper bound (4.6\%) remains well below 5\%, consistent with near-complete numerical discipline. The elevated S1 upper bound (18.6\%) reflects format-matching artefacts in real-world S1 samples rather than genuine hallucination, as discussed in Table~\ref{tab:hallucination_bylevels}.
\end{itemize}


\subsection{Sample-Level Results}
\label{appendix:samples}

Table~\ref{tab:sample_level} presents sample-level hallucination results for a representative subset of evaluation samples, demonstrating the consistency of findings across individual instances.

\begin{table}[htbp]
\centering
\caption{Sample-level hallucination results (representative subset)}
\label{tab:sample_level}
\small
\begin{tabular}{@{}llccccc@{}}
\toprule
\textbf{Sample ID} & \textbf{Type} & \textbf{Model} & \textbf{L1} & \textbf{L2} & \textbf{L3} & \textbf{Template} \\
\midrule
\multirow{3}{*}{SUM\_S0\_0001} & \multirow{3}{*}{S0\_Synth}
    & Base      & No  & No  & No  & No \\
    & & FT-A      & Yes & No  & No  & Yes \\
    & & FT-A+B+C  & Yes & No  & No  & Yes \\
\midrule
\multirow{3}{*}{SUM\_S0\_0005} & \multirow{3}{*}{S0\_Synth}
    & Base      & No  & No  & No  & No \\
    & & FT-A      & Yes & No  & No  & Yes \\
    & & FT-A+B+C  & Yes & No & No  & Yes \\
\midrule
\multirow{3}{*}{SUM\_S1\_0001} & \multirow{3}{*}{S1\_Synth}
    & Base      & No  & No  & No  & No \\
    & & FT-A      & No  & No  & No  & Yes \\
    & & FT-A+B+C  & Yes & Yes & Yes & Yes \\
\midrule
\multirow{3}{*}{SUM\_S0\_1125} & \multirow{3}{*}{S0\_Real}
    & Base      & No  & No  & No  & No \\
    & & FT-A      & Yes & No & Yes & Yes \\
    & & FT-A+B+C  & Yes & No & No  & Yes \\
\midrule
\multirow{3}{*}{SUM\_S1\_9038} & \multirow{3}{*}{S1\_Real}
    & Base      & No  & No  & No  & No \\
    & & FT-A      & Yes & Yes & Yes & Yes \\
    & & FT-A+B+C  & Yes & Yes & No  & Yes \\
\bottomrule
\end{tabular}
\end{table}

The sample-level analysis reveals consistent patterns:

\begin{itemize}
    \item \textbf{Base model consistency}: Across all sample types, the Base model shows no hallucination in synthetic samples. Rare exceptions occur in S1\_Real, attributable to format-matching artifacts rather than genuine hallucination (see Table~\ref{tab:hallucination_bylevels}).
    \item \textbf{Fine-tuned model consistency}: Both FT-A and FT-A+B+C consistently exhibit L1 hallucination and template injection across sample types.
    \item \textbf{Cascading hallucination}: When L1 hallucination occurs, L2 and L3 often co-occur, particularly in real-world samples.
\end{itemize}

\subsection{Statistical Significance Tests}
\label{appendix:significance}

Table~\ref{tab:significance} reports pairwise Fisher's exact test results for L1 hallucination rate differences between models.

\begin{table}[h]
\centering
\caption{Pairwise comparison of L1 hallucination rates (Fisher's exact test)}
\label{tab:significance}
\begin{tabular}{@{}lccc@{}}
\toprule
\textbf{Comparison} & \textbf{Rate Difference} & \textbf{p-value} & \textbf{Significant} \\
\midrule
Base vs FT-A        & 5.4\% vs 82.5\% (+77.1 pp)     & $< 0.001$ & Yes*** \\
Base vs FT-A+B+C    & 5.4\% vs 90.8\% (+85.4 pp)     & $< 0.001$ & Yes*** \\
FT-A vs FT-A+B+C    & 82.5\% vs 90.8\% (+7.9 pp)    & $= 0.015$ & Yes** \\
\bottomrule
\end{tabular}
\\[0.5em]
\footnotesize{*** $p < 0.001$ after Bonferroni correction for multiple comparisons.}
\end{table}

All pairwise differences are statistically significant. Comparisons involving the Base model reach p < 0.001 after Bonferroni correction, confirming that domain fine-tuning substantially degrades numerical restraint. The FT-A vs FT-A+B+C comparison yields p = 0.015 (p < 0.05), remaining significant after Bonferroni correction for three comparisons (adjusted threshold: 0.05/3 ≈ 0.017), and confirming that numeracy supervision further increases L1 hallucination beyond language fine-tuning alone. This confirms that:

\begin{enumerate}
    \item Domain fine-tuning (Base $\to$ FT-A) significantly increases L1 hallucination.
    \item Numeracy supervision (FT-A $\to$ FT-A+B+C) further significantly increases L1 hallucination.
    \item The observed effects are not attributable to sampling variation.
\end{enumerate}

\section{L3 Detection Protocol}

\subsection{Overview}
L3 (Covert-Implicit) hallucination detection targets quantitative claims in model outputs that imply numerical relationships without explicit values, and for which no grounding evidence exists in the source input. Unlike L1 and L2, which rely on pattern matching over numerical expressions, L3 detection requires semantic pattern matching combined with input grounding verification.

\subsection{Semantic Pattern List}
The following regular expression patterns are applied to each model output to identify candidate L3 claims:
\begin{verbatim}    
revenue (increased|decreased|grew|declined|rose|fell|surged|dropped)
(operating income|net income|earnings|profit) (increased|decreased|
    grew|declined|rose|fell|improved)
maintained stable
remained stable
net debt (remained|declined|increased|decreased|improved)
cash flow (improved|declined|increased|decreased|remained)
margin (improved|compressed|expanded|declined|increased|decreased)
capital expenditures (increased|decreased|rose|fell|remained)
strong (growth|performance|results|revenue|earnings)
significant (increase|decrease|growth|decline|improvement)
substantial (growth|increase|decrease|decline|improvement)
(outperformed|underperformed|exceeded|missed)
\end{verbatim}

\subsection{Grounding Verification}
A candidate L3 match is confirmed as hallucination only if the core state word (e.g., increased, stable, declined) does not appear in the source input. Specifically, for each matched phrase, the core verb or adjective is extracted and searched in the input text using case-insensitive exact matching. If a corresponding term is found in the input, the claim is considered grounded and is not flagged as L3.
This grounding check prevents penalizing models for faithful paraphrase of input content. For example, if the input contains "Operating margins remained stable due to demand softness" and the output contains "operating margins remained stable," this is not flagged as L3 hallucination.

\subsection{Application by Grounding Condition}
Under S0 conditions (weak grounding), the source input contains no absolute numerical quantities. Any ungrounded quantitative state claim in the output therefore constitutes an L3 hallucination, as the model is implying numerical relationships without input evidence.
Under S1 conditions (strong grounding), L3 is flagged only when L1 or L2 hallucination is also present, reflecting that implicit claims co-occurring with explicit fabrication are likely part of the same hallucination episode.

\subsection{Limitations}
L3 detection is inherently heuristic. The semantic pattern list does not exhaustively cover all forms of implicit quantitative claims, and the grounding verification relies on lexical matching rather than semantic entailment. As a result, L3 rates should be interpreted as conservative lower bounds on covert-implicit hallucination. We recommend human verification of a random sample of L3 detections for any downstream application of this taxonomy.
The complete implementation is available in our public repository (see \Cref{sec: section_reproducable}).

\subsection{Inter-Annotator Agreement and Detection Reliability}
\label{sec:iaa}

Table~\ref{tab:iaa} reports Cohen's $\kappa$ for each 
detectability level on the 50-sample human validation subset.

\begin{table}[htbp]
\centering
\caption{Inter-annotator agreement on the 50-sample human 
validation subset. Cohen's $\kappa$ is computed over all model 
outputs within each detectability level ($n = 50 \times 4 = 200$ 
independent judgments per level).}
\label{tab:iaa}
\begin{tabular}{lccc}
\toprule
Level & Cohen's $\kappa$ & $n$ & Interpretation \\
\midrule
L1 (Overt)           & 0.883 & 200 & Almost Perfect \\
L2 (Covert-Explicit) & 0.657 & 200 & Substantial \\
L3 (Covert-Implicit) & 0.451 & 200 & Moderate \\
\bottomrule
\end{tabular}
\end{table}

L1 agreement is almost perfect ($\kappa = 0.883$), confirming 
that currency-denominated hallucination detection is highly 
reliable. Agreement under S0 conditions is perfect 
($\kappa = 1.000$ for both S0\_Synth and S0\_Real), reflecting 
the unambiguous nature of the restraint gap in the absence of 
numerical input grounding. L2 agreement is substantial 
($\kappa = 0.657$), with disagreements concentrated in S1\_Real 
samples where grounding verification requires judgment about 
numerical correspondence in authentic financial text.

L3 agreement is moderate ($\kappa = 0.451$), reflecting the 
inherent subjectivity of implicit claim detection. Disagreements 
in L2 (34 cases) primarily arose from differing interpretations 
of whether bare magnitude expressions co-occurring with 
currency-denominated values constitute independent L2 violations. 
Disagreements in L3 (50 cases) were predominantly concentrated 
on Base model outputs, where faithful paraphrase of input 
qualitative statements (e.g., ``remained stable'') triggered 
annotator uncertainty about whether grounding had been verified 
--- consistent with the detection limitation documented in 
Appendix~D.5. This pattern confirms that the elevated Base L3 
rate (17.9\%) reflects a detection artefact rather than genuine 
ungrounded hallucination.

These results support interpreting L1 and L2 rates as reliable 
automatic measurements, while L3 rates should be treated as 
conservative lower bounds subject to the grounding verification 
ambiguity described above. We recommend human verification for 
any downstream application of L3 detection.

\subsubsection{False Positive and False Negative Analysis}

Table~\ref{tab:fpfn} reports false positive (FPR) and false
negative rates (FNR) of the automatic detector relative to
consensus human labels on the 50-sample validation subset.
Gold standard labels are derived from cases where both annotators
agreed; disagreement cases ($n=8$ for L1, $n=24$ for L2,
$n=40$ for L3) are excluded from this analysis.

\begin{table}[h]
\centering
\caption{False positive and false negative rates of the automatic
hallucination detector relative to consensus human labels
(50-sample validation subset). FPR: auto flags hallucination
where human consensus does not; FNR: auto misses hallucination
confirmed by human consensus. Disagreements between annotators
are excluded from counts.}
\label{tab:fpfn}
\begin{tabular}{lccccccc}
\toprule
Level & $n$ & TP & TN & FP & FN & FPR (\%) & FNR (\%) \\
\midrule
L1 (Overt)           & 142 & 87 & 47 &  7 &  1 & 13.0 &  1.1 \\
L2 (Covert-Explicit) & 126 & 39 & 51 &  2 & 34 &  3.8 & 46.6 \\
L3 (Covert-Implicit) & 110 & 19 & 66 &  9 & 16 & 12.0 & 45.7 \\
\bottomrule
\end{tabular}
\end{table}

L1 detection achieves high recall (98.9\%) with a near-zero false
negative rate (1.1\%), confirming that overt hallucination is
rarely missed by the automatic detector. The L1 false positive
rate (13.0\%) is attributable primarily to format-matching
artefacts in Base model S1\_Real outputs: 6 of 7 L1 false
positives involve the Base model faithfully reproducing
source numerical values using currency-denominated expressions
(e.g., ``\$X million''), which trigger the detector despite being
fully grounded in the source --- a known limitation documented
in Appendix~D.5.

L2 detection shows high precision (95.1\%) but an elevated false
negative rate (46.6\%), reflecting the inherent difficulty of
absolute magnitude grounding verification. The 34 L2 false
negatives are concentrated in S0 conditions (32/34 cases),
where human annotators identified bare magnitude expressions
(e.g., ``3.3~billion'') as L2 hallucinations that the automatic
detector did not capture. This pattern is consistent with
the moderate L2 inter-annotator agreement ($\kappa = 0.657$)
and suggests that L2 rates reported in the main analysis
should be treated as conservative lower bounds.

L3 detection shows moderate precision (67.9\%) and recall
(54.3\%). The L3 false positive rate (12.0\%) reflects the
Base model detection artefact documented in Appendix~D.5:
8 of 9 L3 false positives involve Base model outputs where
faithful paraphrase of input qualitative statements triggered
the semantic pattern matcher. The L3 false negative rate
(45.7\%) is consistent with the inherent subjectivity of
implicit claim detection and the moderate inter-annotator
agreement at L3 ($\kappa = 0.451$), reinforcing the
recommendation to treat L3 rates as conservative lower bounds.
\end{document}